%% file: main.tex
\documentclass[sigconf]{acmart}

\usepackage{multicol,url,color}
\usepackage{multirow}
\usepackage{rotating,booktabs}
\usepackage{adjustbox,subcaption}
\usepackage{amsmath,amsfonts}
\usepackage{algorithm}
\usepackage[noend]{algpseudocode}
\usepackage{graphicx}
\usepackage[multiple]{footmisc}
\usepackage{caption}
\usepackage{bbding}
\usepackage{colortbl}
\usepackage{hhline}
\usepackage{multirow}
\usepackage{comment}
\usepackage{textcomp}
\usepackage{pifont}
\makeatother
\usepackage{float}
\usepackage{fancyhdr}
\usepackage{mathtools} 
\usepackage{graphics}
\usepackage{placeins}
\usepackage{array}
\usepackage{enumitem}
 \usepackage{color}

 \usepackage{tikz}
  \makeatletter

\newcommand{\vect}[1]{\mathbf{#1}}
\newcommand{\Wskip}[1]{ }

\renewcommand\fbox{\fcolorbox{gray}{white}}

\copyrightyear{2022} 
\acmYear{2022} 
\setcopyright{acmcopyright}\acmConference[WWW '22 Companion]{Companion Proceedings of the Web Conference 2022}{April 25--29, 2022}{Virtual Event, Lyon, France}
\acmBooktitle{Companion Proceedings of the Web Conference 2022 (WWW '22 Companion), April 25--29, 2022, Virtual Event, Lyon, France}
\acmPrice{15.00}
\acmDOI{10.1145/3487553.3524713}
\acmISBN{978-1-4503-9130-6/22/04}

\begin{document}

\sloppy
\title{MarkovGNN: Graph Neural Networks on Markov Diffusion}

\author{Md. Khaledur Rahman}
\affiliation{%
  \institution{Indiana University Bloomington}
  \country{}}
\email{morahma@iu.edu}
\author{Abhigya Agrawal}
\affiliation{%
  \institution{Indiana University Bloomington}
  \country{}}
\email{abagra@iu.edu}
\author{Ariful Azad}
\affiliation{%
 \institution{Indiana University Bloomington}
  \country{}}
\email{azad@iu.edu}

\begin{abstract}
Most real-world networks contain well-defined community structures where nodes are densely connected internally within communities. 
To learn from these networks, we develop MarkovGNN that captures the formation and evolution of communities directly in different convolutional layers.
Unlike most Graph Neural Networks (GNNs) that consider a static graph at every layer, MarkovGNN generates different stochastic matrices using a Markov process and then uses these community-capturing matrices in different layers.
MarkovGNN is a general approach that could be used with most existing GNNs.
We experimentally show that MarkovGNN outperforms other GNNs for clustering, node classification, and visualization tasks. The source code of MarkovGNN is publicly available at \url{https://github.com/HipGraph/MarkovGNN}.
\end{abstract}


\maketitle
\input{intro}

\input{methods}

\input{results}

\section{Conclusions}

Communities in a network can be discovered by probabilistic random walks with pruning that iteratively promote intra-cluster edges and demote inter-cluster edges. 
The proposed algorithm MarkovGNN captures snapshots of this community discovery process in different layers of graph neural networks. 
We conclude with experimental evidence that using Markov matrices in different layers improve the performance of clustering, node classification and visualization tasks.  
MarkovGNN excels on graphs having well-defined community structures (e.g., a high value of the clustering coefficient) because it can preserve the communities in the embedding space. 
However, if a graph lacks a community structure, MarkovGNN performs no worse than GCN.
Overall, MarkovGNN brings in the philosophy of using different community-capturing graphs at different layers of a GNN and provides a flexible design space for graph representation learning. 


\begin{acks}
This research was supported by the Applied Mathematics Program of the DOE Office of Advanced Scientific Computing Research under contract number DE-SC0022098.
\end{acks}

\bibliographystyle{ACM-Reference-Format}
\bibliography{main}

\appendix
\input{supplementary}

\end{document}

%% file: intro.tex
\vspace{-2pt}
\section{Introduction}
\vspace{-2pt}

Graph representation learning using graph neural networks (GNNs) is often formulated by a message passing model~\cite{gilmer2017neural}.
In this model, a node $u$ receives messages from its direct neighbors and then updates $u$'s embedding based on the received messages and its own features.
While a GNN with $L$ layers indirectly exchanges messages among $L$-hop neighbors, messages in each layer are still restricted to direct neighbors.
Message passing via direct neighbors 
on the same static graph structure may limit the expressive power of GNNs. 
Can we address this limitation by generating a  series of graphs (using a diffusion process) created from a given static structure and then using the generated graphs at different layers of GNN?
In this paper, we answer this question affirmatively using a Markov diffusion process~\cite{Dongen2000} to generate a series of graphs.
The goal is not to develop a new GNN method, but to use Markov diffusion to improve the performance of any existing GNN.
When an existing GNN method uses different Markov matrices at different layers, we call this augmented GNN a MarkovGNN.



\begin{figure}
    \centering
    \includegraphics[width=\linewidth]{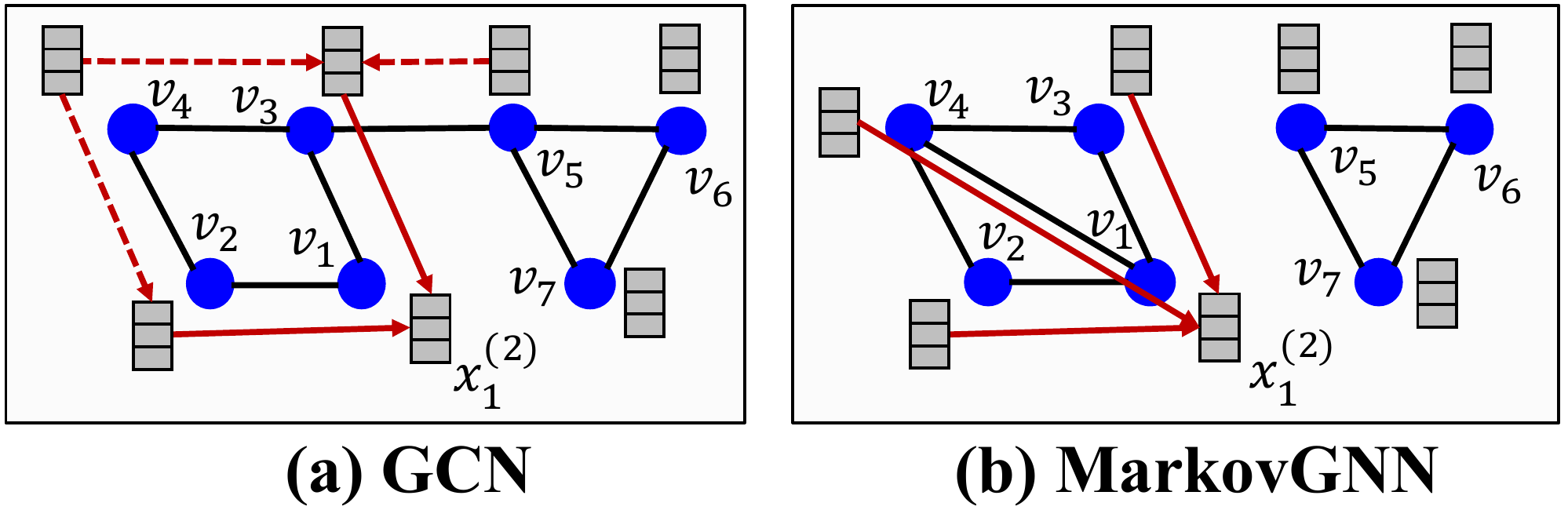}
    \vspace{-18pt}
    \caption{Updating the embedding $x_1^{(2)}$ of $v_1$ at the second layer of GNN. (a) For GCN, $x_1^{(2)}$ is computed directly from the embeddings of $v_2$ and $v_3$ (shown by solid arrowheads), which indirectly depend on the embeddings of $v_1$'s two-hop neighbors $v_4$ and $v_5$ (shown by dashed arrowheads). (b) The second layer of MarkovGNN uses a different graph structure with an edge $\{v_1,v_4\}$ inserted and an edge $\{v_3,v_5\}$ deleted possibly because $v_4$ is in the same community of $v_1$ whereas $v_5$ is in a different community.
    This graph structure is updated based on a Markov process that captures the community pattern in the graph. Thus, in MarkovGNN, $x_1^{(2)}$ is computed directly from the embeddings of $v_2$, $v_3$, and $v_3$ (shown by solid arrowheads), which are neighbors of $v_1$ in the current graph.}
    \label{fig:messages}
    \vspace{-0.5cm}
\end{figure}

The central idea of MarkovGNN is to use a series of graphs at different layers of a GNN so that different layers do not operate on the same static graph structure. 
To understand the limitation of using the same static graph at every layer, Fig.~\ref{fig:messages}(a) considers the message passing at the second layer of  a graph convolutional network (GCN)~\cite{kipf2016semi}.
At the second layer of GCN, node $v_1$ receives information from 2-hop neighbors ($v_4$ and $v_5$) indirectly via 1-hop neighbors ($v_2$ and $v_3$).
Such indirect messages dilute information received from high-order neighbors and may create the  ``over-smoothing"~\cite{alon2020bottleneck, oono2019graph} problem resulting from hidden-layer representations being squeezed into fixed-length vectors.
What if we eliminate indirect messaging by using a new structure created from the original graph? 
Fig.~\ref{fig:messages}(b) shows a modified graph where we added a new edge $\{v_1, v_4\}$ and deleted an existing edge $\{v_3, v_5\}$ based on a diffusion model (here, to capture the community structure in the original graph). 
If we use the modified graph shown in Fig.~\ref{fig:messages}(b) at the second layer of a GNN, we promote direct messages (from $v_4$) and  suppress indirect messages (from $v_5$). 
We argue that using a series of modified graphs in the hidden layers for direct messaging instead of a static graph for indirect messaging benefits most existing GNN methods.

There are many ways to perturb a given graph to generate a series of graphs such as the heat kernel and personalized PageRank. 
In this paper, we use a modified Markov diffusion process~\cite{van2001graph} because if its flexibility in controlling diffusion while capturing the community structures in the graph. 
The Markov process generates a sequence of graphs (called Markov matrices) with an aim to find communities or clusters in the original graph (a community represents a subset of nodes that are densely connected to each other and loosely connected to other communities). 
Thus, the sequence of Markov matrices promotes or demotes edges based on their relevance to the underlying clustering pattern. 
Consequently, using Markov matrices at different layers of a GNN helps learning from graphs that expose  homophily~\cite{zhu2020beyond} by organizing nodes into communities.

For graphs containing clustering patterns (e.g., graphs with high clustering coefficients measured by counting the number of triangles in the graph), MarkovGNN 
offers the following benefits.
(1) By simulating the community formation process, MarkovGNN directly captures community patterns into different layers of a GNN.
(2) Different Markov matrices 
increasingly 
put more weight on intra-cluster edges and less weight on inter-cluster edges. Thus, the Markov process (not the GNN) automatically controls information flows on paths based on their relevance to form communities.   
This {\em community-driven flow} eliminates the need to sample neighbors  used in GraphSAGE \cite{hamilton2017inductive} or sample graphs used in  GraphSAINT \cite{zeng2019graphsaint}. 
(3) As shown in Fig.~\ref{fig:messages} (b), MarkovGNN only exchanges messages among 1-hop neighbors on modified graphs. Hence, it can prevent  the ``over-smoothing"~\cite{alon2020bottleneck, oono2019graph} problem at high-degree nodes.
(4) 
By pruning low probable diffusion paths, the Markov process keeps the graph sparse at every layer of the GNN and prevents the ``neighbor-explosion"~\cite{hamilton2017inductive} problem.
(5) MarkovGNN is more expressive than Cluster-GCN~\cite{chiang2019cluster} that precomputes clusters and uses the clustering at every layer of GCN. We will show that a special variant of MarkovGNN is equivalent to Cluster-GCN.


This paper makes the following key contributions. 
\begin{enumerate}[topsep=3pt, itemsep=0pt, leftmargin=15pt]
\item We present MarkovGNN that uses a series of community-aware sparse adjacency matrices at different layers of a GNN. 
MarkovGNN can augment most GNN methods, and it is more general than GDC~\cite{klicpera2019diffusion} and Cluster-GCN~\cite{chiang2019cluster}.

\item By using residual connections, MarkovGNN supports deeper GNNs. 
MarkovGNN reduces the neighbor-explosion problem in deeper layers by keeping Markov matrices sparse.



\item We conduct an extensive set of experiments to evaluate the performance of MarkovGNN with respect to baseline GNN methods that use the same graph at every layer. If vertices in the original graph are well-clustered, MarkovGNN preserves the clustering pattern in the embedding space and improves the performance of clustering, node classification, and visualization tasks.
\end{enumerate}



%% file: methods.tex
\vspace{-5pt}
\section{Background and Motivation}
\vspace{-3pt}
Let $G(V,E)$ denote a graph with $n=|V|$ vertices and $|E|$ edges.
The adjacency matrix is denoted by an $n\times n$ sparse matrix $\vect{A}$ where $\vect{A}[i,j] = 1$ if $\{v_i,v_j\}{\in}E$, otherwise $\vect{A}[i,j] = 0$.
Additionally, $\vect{X}\in \mathbb{R}^{n\times f}$ denotes the feature matrix where the $i$th row stores $f$-dimensional features for the $i$th vertex.  
The $l$th convolution layer of a GNN computes embeddings of every node based on the embeddings of neighbors as follows: 
\vspace{-3pt}
\begin{equation}
\label{eq:GCN}
   \vect{X}^{(l+1)} = \sigma(\vect{A}\vect{X}^{(l)}\vect{W}^{(l)}), 
\end{equation}
where $\vect{X}^{(l)}{\in} \mathbb{R}^{n\times f_l}$ is $f_l$-dimensional embeddings used as input to the $l$th layer, $\vect{X}^{(l+1)}{\in} \mathbb{R}^{n\times f_{l+1}}$ is $f_{l+1}$-dimensional embeddings generated as the output from the $l$th layer and $\vect{W}^{(l)}{\in} \mathbb{R}^{f_{l} \times f_{l+1}}$ is the weight matrix.
In most instances of GNN, the adjacency matrix is normalized and ReLU is used as the activation function. 




{\bf The limitation of GNNs that use a fixed graph at every layer.}
If a static graph is used at every layer of a GNN, it  only passes messages between adjacent nodes where the adjacency remains fixed throughout the execution of the algorithm. 
While GCN and other message passing networks do leverage higher-order neighborhoods in deeper layers, all messages must pass through immediate neighbors which create bottlenecks when flowing messages from higher-order neighbors~\cite{alon2020bottleneck}.
To remove such bottlenecks, high-pass and low-pass filters were proposed in a recent work with an aim to control the information flow~\cite{bo2021beyond}.
However, in almost all GNNs, the graph structure remains static at different layers.
We argue that distinct graph structures used in different layers of GNNs could benefit most existing GNN methods. 
In particular, we aim to capture the formation of communities in different layers of GNN to improve the expressivity of GNN models.  

{\bf Previous work that uses diffusion in GNNs.} The most notable approach is called graph diffusion
convolution (GDC)~\cite{klicpera2019diffusion} where a diffusion matrix $\Tilde{\vect{A}}$ is constructed via a generalized graph diffusion process: 
 \vspace{-12pt}
\begin{equation}
    \label{eq:gdc}
    \Tilde{\vect{A}} = \sum_{j=0}^{\infty} \beta_j \vect{T}^j, 
    \vspace{-0.17cm}
\end{equation}
where $\beta_j$ is the weighting coefficient and $\vect{T}$ is a column stochastic adjacency matrix computed by $\vect{A}\vect{D}^{-1}$ and $\vect{D}$ is the diagonal matrix of node degrees.
The idea used in GDC is interesting because of the well-known fact that graph diffusion can act as a denoising filter similar to Gaussian filters on images~\cite{egilmez2018graph}.
However, GDC creates the diffusion matrix as a pre-processing step and uses the same diffusion matrix $\Tilde{\vect{A}}$ in every layer of a GNN.
Hence, GDC still uses the same connectivity matrix (albeit different from the original graph) in every layer and thus misses the opportunity to incrementally learn from the diffusion process. 
Furthermore, the diffusion process typically results in a dense matrix, making the computation  prohibitively expensive. 
Even though GDC sparsifies $\Tilde{\vect{A}}$ before using it in GNNs, the computation of $\vect{T}^j$ becomes the bottleneck.
Our goal is to capture the diffusion process (more specifically via a Markov process) directly in different layers of GNN.
We show that our algorithm performs much better than GDC in identifying clustering patterns and classifying nodes. 

\textbf{Related work.} 
Graph representation learning is a well-studied research area with numerous unsupervised and semi-supervised methods proposed in the literature. 
Early unsupervised works focused on graph embedding based on neighborhood exploration such as random-walk based methods \cite{perozzi2014deepwalk,grover2016node2vec,ribeiro2017struc2vec,rahman2020force2vec} and $k$-hop neighborhood exploration-based methods \cite{tang2015line}. 
While unsupervised embedding methods can capture the structure of a graph, they often cannot utilize node features or domain-specific labels.
Graph Convolution Networks (GCN) is one of the first semi-supervised methods which used graph convolutions for better  representation learning \cite{kipf2016semi}. Later, many other GNN methods have been proposed to tackle evolving issues in GNNs and improve graph learning, e.g., GraphSAGE to incorporate different types of aggregation strategies \cite{hamilton2017inductive}, Graph Attention Networks (GAT) to consider the asymmetric influence of neighbors \cite{velivckovic2017graph,abu2017watch}, FastGCN to improve the training process \cite{chen2018fastgcn}, Cluster-GCN to resolve neighborhood expansion problem and train deep networks \cite{chiang2019cluster}. Some recent efforts pre-process the original graph and/or input features to create a modified graph feeding it to all convolutional layers of the GNN \cite{klicpera2019diffusion}. 
We refer readers to a comprehensive survey for further details~\cite{wu2020comprehensive}. 
Most GNN methods use a static graph (often preprocessed) in all GNN layers \cite{kipf2016semi,hamilton2017inductive,klicpera2019diffusion,yang2020factorizable,suresh2021breaking} whereas some methods use a mixing contribution of neighbors~\cite{abu2019mixhop,wu2019simplifying,liu2020towards}. 
GraphSAINT uses different sampled subgraphs in different minibatches but it still uses the same subgraph at every layer \cite{zeng2019graphsaint}. Different from previous work, we put efforts to use different graph structures in different convolutional layers. 

\section{The Markov diffusion process}
\vspace{-2pt}
In this section, we review the Markov process~\cite{Dongen2000} that simulates flow within a graph such that the flow eventually gets trapped in communities at the convergence of the process. 
The flow is simulated by random walks with the assumption that higher-length paths between two nodes within a community should stay within the community.
The Markov process controls random walks by computing transition  probabilities between every pair of vertices.
The process incrementally promotes the probability of intra-cluster walks, demotes the probability of inter-cluster walks, and removes paths with low probabilities to discover natural clusters in the graph.
Markov process consists of four operations: (a) starting with a column (or row) stochastic transition matrix and maintaining it until convergence, (b) expanding random walks by the {\bf expansion} operation, (c) strengthening intra-cluster walks and weakening inter-cluster walks by the {\bf inflation} operation, and (d) removes paths with low probabilities by the {\bf prune} operation.
These operations are repeated until converges.

{\bf The transition matrix.} Let $\vect{D}$ be a diagonal matrix with $\vect{D}[i,i]=\sum_{j=1}^{n} \vect{A}[i,j]$, where $\vect{A}$ is the (weighted) adjacency matrix of the graph.
The Markov process starts with a column-stochastic matrix $\vect{M}=\vect{A}\vect{D}^{-1}$.
The $i$th column of $\vect{M}$ sums to one, and $\vect{M}[i,j]$ is interpreted as the probability of following the edge from vertex $j$ to vertex $i$ in the random walk.

{\bf The expansion operation.}
The Markov process expands random walks by squaring the Markov matrix $\vect{M}$.
Thus, the expansion operation in the $i$th iteration computes $\vect{M_{i-1}}{\times} \vect{M_{i-1}}$ where $\vect{M_{i-1}}$ is the output of the previous iteration. 
The expansion step is similar to the generalized graph
diffusion~\cite{klicpera2019diffusion} and typically converges to an idempotent matrix according to the Perron–Frobenius theorem. 

{\bf The inflation operation.}
The parameterized inflation operation with parameter $r$ performs 
Hadamard power of the matrix denoted by $\vect{M}^{\circ r}$ and makes the matrix column stochastic again as follows:
\vspace{-10pt}
\begin{equation}
  \vect{M}^{\circ r}[i,j] = \vect{M}[i,j]^{r}/\sum_{i=1}^{n}\vect{M}[i,j]^{r}
\end{equation}
The goal of the non-linear inflation step is to strengthen higher probability paths (intra-cluster connections) and weaken lower probability paths (inter-cluster connections). 

{\bf The prune operation.}
To keep the Markov matrices sparse, we prune small entries that are below a threshold $\theta$.
This step ensures that random-paths are contained within communities and that the matrix does not become dense.
\setlength{\textfloatsep}{4pt}
\begin{algorithm}[!tb]
\caption{Generating Markov matrices}\label{algo:mcl}
\begin{flushleft}
\textbf{Input:} $\vect{A}$: adjacency matrix of the graph, $r:$ value of element-wise inflation, $\theta:$ pruning threshold\\
\textbf{Output:} $\vect{M}_1, \vect{M}_2,..., \vect{M}_k$: column stochastic matrices from all steps of the Markov process.  
\end{flushleft}
\begin{algorithmic}[1]
\Procedure{Markov}{$\vect{A}, r, \theta$}
    \State $\vect{M_1} \gets \text{ColStochastic}(\vect{A})$ 
    \For{$k\gets 2..$}
    \State $\vect{M}_k \gets  \vect{M}_{k-1}\times \vect{M}_{k-1}$ \Comment{the expansion operation}
    \State $\vect{M}_k \gets \text{ColStochastic}(\vect{M}_k^{\circ r})$\Comment{inflation}
    \State $\vect{M}_k \gets \text{ColStochastic}(\text{Prune}(\vect{M}_k, \theta))$
    \Comment{Sparsify the matrix by pruning small
entries less than $\theta$}
    \If{$\vect{M}_k\approx \vect{M}_{k-1}$}
    \State \Return $\vect{M}_1, \vect{M}_2,..., \vect{M}_k$
    \EndIf
    \EndFor
\EndProcedure
\end{algorithmic}
\end{algorithm}

{\bf Generating Markov matrices for GNNs.}
Algorithm~\ref{algo:mcl} shows the steps used to generate Markov diffusion matrices.
The algorithm starts with a column stochastic matrix $\vect{M}_1$ where each column sums to 1 and maintains column stochastic matrices in each iteration. 
The $k$th iteration performs the expansion, inflation, and  pruning operations.
The algorithm converges when the stochastic matrix does not change in successive iterations.
We store all stochastic matrices $\vect{M}_1, \vect{M}_2,..., \vect{M}_k$ from different iterations so that they can be used in different layers of a GNN.
While there is a memory overhead to store $k$ Markov matrices, it makes the GNN training much faster as we will show in the next section.   

\begin{figure*}[!t]
    \centering
    \vspace{-8pt}
    \includegraphics[width=.97\linewidth]{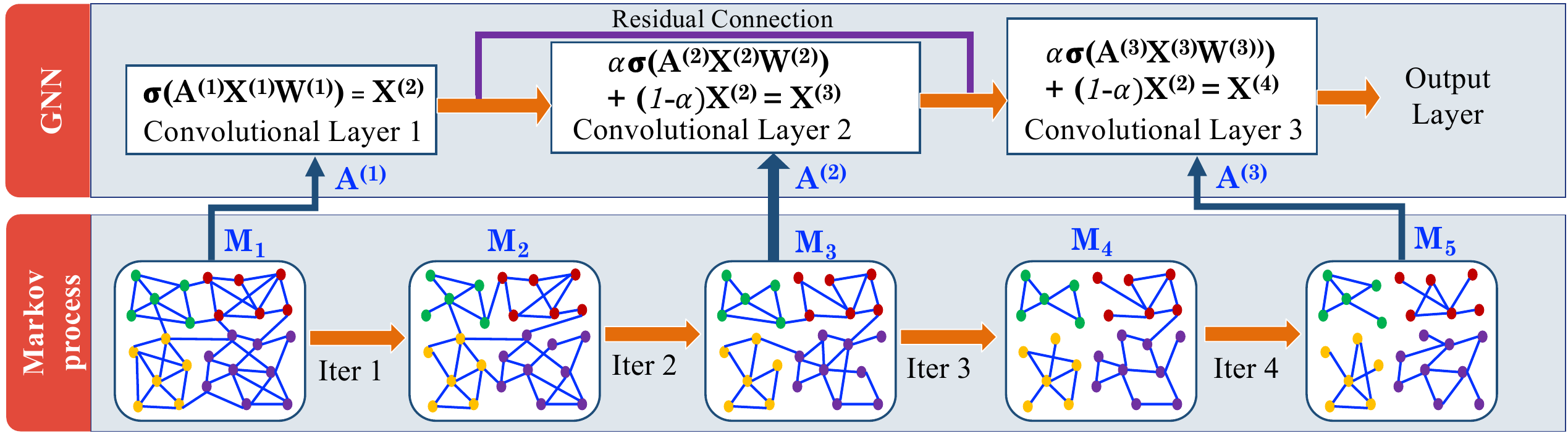}
    \vspace{-10pt}
    \caption{The bottom panel shows column stochastic matrices generated by the Markov process for hypothetical graphs with bi-directional edges shown by lines. Here, $\vect{M}_1$ is the input stochastic matrix representing the given graph, $\vect{M}_5$ is the converged matrix with four discovered communities, and $\vect{M}_2$, $\vect{M}_3$, and $\vect{M}_4$ are intermediate matrices capturing the formation of communities. As communities are discovered, edges are added and deleted in different iterations. 
    The top panel shows a GNN that uses a subset of Markov matrices in three convolutional layers where $\alpha$ is the contribution factor of residual connections.
    }
    \vspace{-10pt}
    \label{fig:introduction}
\end{figure*}

{\bf Convergence.} It has already been shown that the Markov process converges quadratically~\cite{van2001graph}. In most practical graphs, the process converges in less than 30 iterations. 
Upon convergence, connected components are treated as clusters in the original graph. 



{\bf Hyperparameters in the Markov process.} The exponent $r$ in inflation and the pruning threshold $\theta$ are hyperparameters that could be tuned to obtained different matrices. 
A higher value of the parameter $r$ demotes flow along long paths, which creates fine-grained communities. 
Thus, the number and granularity of final communities can be controlled by $r$. When $r$ is set to 1 (i.e., no inflation), the Markov process is equivalent to a general diffusion process~\cite{klicpera2019diffusion}.
$\theta$ controls the sparsity of Markov matrices. 
We empirically identify suitable values of these parameters.

\vspace{-6pt}
\section{The {MarkovGNN} Algorithm}
\label{sec:proposedalgorithm}
\vspace{-2pt}
The MarkovGNN algorithm exchanges the static adjacency matrix with a series of Markov matrices at different layers of a GNN. 
This approach could be used with most popular GNN methods such as GCN, GAT, and GraphSAINT. 
Figure~\ref{fig:introduction} illustrates the forward propagation of a MarkovGNN.
In this example, the GNN has three layers (shown in the top panel), but the Markov process converges in four iterations (in the bottom panel).
Hence, layer 1, 2, and 3 of GNN use $\vect{M}_1$, $\vect{M}_3$, and $\vect{M}_5$, respectively. 
\begin{algorithm}[!t]
\caption{MarkovGCN forward propagation}\label{algo:mgcn}
\begin{flushleft}
\textbf{Input:} $\vect{A}$: adjacency matrix of an input graph, $\vect{X}$: input feature, $L:$ number of layers, $r$: inflation parameter, $\theta$: pruning threshold\\
\textbf{Output:} $\vect{X}^{(L+1)}$: node representation from the last layer.
\end{flushleft}
\begin{algorithmic}[1]
\Procedure{MarkovGCN}{$\vect{A}, L, r, \theta$}
    \State $\{\vect{M}_1, \vect{M}_2,..., \vect{M}_k \}\gets \text{MARKOV}(\vect{A}, r, \theta)$ \Comment{ Alg.~\ref{algo:mcl}}
    \State $\vect{X}^{(1)} \gets \vect{X}$
    \State $\vect{W}^{(1)},...\vect{W}^{(L)} $ are weight matrices
    \State $\{\vect{A}^{(1)}, \vect{A}^{(2)},..., \vect{A}^{(L)} \}\gets $ select $L$ matrices from $\{\vect{M}_1, \vect{M}_2,..., \vect{M}_k\}$ one for each layer \Comment{assuming $L{<=}k$}
    \For{$l=1$ to $L$}
        \State $\vect{X}^{(l+1)} = \alpha \sigma(\vect{A}^{(l)}\vect{X}^{(l)}\vect{W}^{(l)}) + (1-\alpha) \phi (\vect{X}_1|
        \ldots|\vect{X}_{l-1})$
    \EndFor
\State \Return $\vect{X}^{(L+1)}$
\EndProcedure
\end{algorithmic}
\end{algorithm}

{\bf Residual connections.}
A MarkovGNN needs more layers than a typical GNN to utilize various Markov matrices at different layers. 
To train such deep GNN models, we use residual connection~\cite{he2016deep}  which has also been used in other deep networks~\cite{li2021residual,chen2020simple}. With residual connections, the embedding in the $l$-th layer of MarkovGNN is computed as follows:
\vspace{-2pt}
\begin{equation}
\label{eqn:residual}
    \vect{X}^{(l+1)} = \alpha \sigma(\vect{A}^{(l)}\vect{X}^{(l)}\vect{W}^{(l)}) + (1-\alpha) \phi (\vect{X}_1|
        \ldots|\vect{X}_{l-1})
\end{equation}
\vspace{-2pt}
where, $\sigma$ is the activation function, $\phi$ is a selector and $\alpha$ is a hyper-parameter that controls the contribution of the residual connections. Note that the operator $\phi$ in $l$-th layer selects only one representation matrix from the previous $l-1$ layers.

{\bf The MarkovGNN algorithm.}
Algorithm~\ref{algo:mgcn} describes the forward propagation of the MarkovGCN algorithm.
If a GNN has $L$ layers, we select $L$ matrices $\{\vect{A}^{(1)}, \vect{A}^{(2)},..., \vect{A}^{(L)} \}$ from the collection of stochastic matrices returned by Algorithm~\ref{algo:mcl}.
Then, in the $l$th layer of the GNN, convolution is performed on $\vect{A}^{(l)}$ instead of the original adjacency matrix (line 7 of Algorithm~\ref{algo:mgcn}).
Here, we assume that the number of iterations $k$ needed for the convergence of the Markov process is greater than $L$.
This is a reasonable assumption since $k$ is typically more than 20 for most graphs we considered, and most GNNs have less than 10 layers in practice. 
The selection of $L$ matrices from $k$ available matrices is an algorithmic choice for which we show various experimental results. 
However, we always present Markov matrices to GNN  layers in the same order they are generated in the Markov process. 

{\bf Training MarkovGNN.} In Algorithm \ref{algo:mgcn}, $\sigma$ represents an activation function in line 7 for which either ReLU or LeakyReLU is a common choice. We add a tunable dropout parameter to each convolutional layer. In addition, a softmax layer is added to the last layer of MarkovGNN. Similar to other GNNs, our goal is to learn weight matrices by optimizing a loss function as follows:
\vspace{-4pt}
\begin{equation}
    \mathcal{L} = \sum_{i\in \mathrm{Y}_L} loss(y_i, \vect{X}^{(L)}_i),
\end{equation}
where, $\mathrm{Y}_L$ represents all vertices that have ground truth labels, and $\vect{X}^{(L)}_i$ represents $i$th row of $\vect{X}^{L}$ having ground truth label $y_i$. In MarkovGNN, we use the standard negative log-likelihood loss function and the Adam optimizer \cite{kingma2014adam} to train models.


\subsection{Variants of MarkovGNN}
\vspace{-3pt}
MarkovGNN is a general technique that could be used with any GNN.
For example, when Markov matrices are used with GCN, we call this combination MarkovGCN (similarly, MarkovGAT, MarkovGraphSAINT, etc.).
Furthermore, the Markov matrices can be combined differently to obtain several interesting variants of GNNs as shown in Table~\ref{tab:markov_variants} and described below.

\begin{table}
\centering
  \caption{GNN variants by combining Markov matrices.\vspace{-7pt}}
    \begin{tabular}{l l l |c }
        \toprule
        $\vect{A}^{(1)}$ & $\vect{A}^{(2)}$... & $\vect{A}^{(L)}$ & Equivalent GNN models\\
        \midrule
        $\vect{M}_1$ & $\vect{M}_1$ & $\vect{M}_1$ & GCN~\cite{kipf2016semi}\\
        $\sum_{i=1}^k \vect{M}_i$ & $\sum_{i=1}^k \vect{M}_i$ & $\sum_{i=1}^k \vect{M}_i$ & GDC~\cite{klicpera2019diffusion}\\
        $\vect{M}_k$ & $\vect{M}_k$ & $\vect{M}_k$ &  Cluster-GCN~\cite{chiang2019cluster}\\
        
        $\vect{M}_1$ & $\vect{M}_2$ & $\vect{M}_k$ & General Markov-GNN\\
        \bottomrule
    \end{tabular}
    \label{tab:markov_variants}
\end{table}
{\bf uMarkovGNN: union of Markov matrices.}
In row 2 of Table~\ref{tab:markov_variants} we 
use $\vect{\Tilde{M}}{=}\sum_{i=1}^k \vect{M}_i$ as a unified Markov matrix and use it in all layers of a GNN. 
For example, the forward propagation at layer $l$ of GCN could be computed by 
$\vect{\Tilde{M}}\vect{X}^{(l)}\vect{W}^{(l)}$ where the graph structure $\vect{\Tilde{M}}$ does not change in different layers. 
We call this approach uMarkovGNN which is 
equivalent to the diffusion matrix considered in Eq.~\ref{eq:gdc}.
However, uMarkovGCN may perform better than GDC because the former preserves a unified view of communities. 

{\bf MarkovClusterGNN.} 
In row 3 of Table~\ref{tab:markov_variants} we 
use the converged Markov matrix $\vect{M}_k$ at every layer of GNN (that is, the computation at layer $l$  is $\vect{M}_k\vect{X}^{(l)}\vect{W}^{(l)}$). We call this approach MarkovClusterGNN, which is equivalent to Cluster-GCN~\cite{chiang2019cluster}.
Cluster-GCN also identifies clusters using a graph partitioner and then uses the clustered graph at every layer of GCN.  



\vspace{-6pt}
\subsection{Properties of MarkovGNN}
\vspace{-3pt}
{\bf Computational complexity.}
Let $\vect{M}_k$ have $d_k$ nonzeros per column (that is, the average degree of the corresponding graph is $d_k$).
Then, the expected cost of computing   $\vect{M}_k$ in line 4 of Algorithm~\ref{algo:mcl} is $nd_k^2$~\cite{ballard2013communication}.
Similarly, let $\vect{A}^{(l)}$ have $d_l$ nonzeros per column on average. 
Then the computational cost of the forward propagation is 
\vspace{-5pt}
\begin{equation}
    \label{eq:complexity}
    \sum_{l=1}^{L} (nd_lf_l + nf_lf_{l+1}).
    \vspace{-6pt}
\end{equation}
{\bf The role of sparsity.} As communities are formed in the Markov process, the corresponding matrices become increasingly sparse. 
As a result, $d_{l+1}$ is typically smaller than $d_l$ in deeper layers of MarkovGNN.
Thus, 
MarkovGNN is expected to run faster than other other diffusion based methods such as GDC.

{\bf An upper bound on the number of layers $L$.}
In most GNN methods, the number of layers is selected in an ad hoc fashion.
By contrast, MarkovGNN sets an upper bound on the number of layers to the number of Markov matrices ($k$) returned by Algorithm~\ref{algo:mcl}.

{\bf A potential solution to the neighbor-explosion problem.}
As communities are formed, the Markov process removes intra-cluster edges via the inflation and pruning operations. 
These steps keep the graph sparse at every layer  and could prevent the neighbor-explosion problem observed in GNNs.

{\bf The role of sampling from $\{\vect{M}_1, \vect{M}_2,..., \vect{M}_k \}$ in GNN layers.}
The selection of $L$ matrices from $k$ available Markov matrices is an algorithmic choice for which we can consider various selection strategies. 
This selection of matrices from a pool of available matrices  makes it unnecessary to sample neighbors  used in GraphSAGE \cite{hamilton2017inductive} or to sample minibatches used in  GraphSAINT \cite{zeng2019graphsaint}.

{\bf Limitations of MarkovGNN.} 
Since the Markov process iteratively identifies the community structure in the graph, MarkovGNN is expected to excel when the graph has well-defined communities. Our experimental results show clear evidence in favor of this hypothesis. 
However, when a graph has no clear clustering patterns 
MarkovGNN may not perform better than other GNN methods.



%% file: results.tex
\vspace{-7pt}
\section{Experiments}
\vspace{-3pt}
\subsection{Experimental Setup}
\vspace{-3pt}
\textbf{Overview of experiments.}
We conduct an extensive set of experiments by coupling Markov matrices with GCN, GAT, and GraphSAINT. 
As MarkovGNN is expected to capture the clustering patterns in a graph, we compare the clustering effectiveness of MarkovGNN with other GNN and unsupervised algorithms.  
We also examine the effectiveness of MarkovGNN in node classification and graph visualization tasks. 


\textbf{Experimental platforms.}
We conduct all experiments on  an Intel Skylake processor  with 48 cores and 128GB memory. We have implemented MarkovGNN using PyTorch Geometric (PyG)~\cite{fey2019fast} version 1.6.1 built on top of PyTorch 1.7.0. We used GCN, GAT, Cluster-GCN, GDC, and GraphSAINT implementations from PyG. We run all experiments 10 times and report the average results.

\begin{table}[!tb]
\centering
\caption{Summary of Datasets. `-' indicates the absence of input features.  Graphs are sorted based on the descending order of the Average Clustering Coefficient (ACC).}
\vspace{-8pt}
\begin{tabular}{c|c|c|c|c|c} 
\hline
\textbf{Graphs} & \textbf{Nodes} & \textbf{Edges} & \textbf{Classes} & \textbf{Features} & \textbf{ACC}  \\ 
\hline
USAir           & 1,190           & 13,599          & 4                & - &     0.50            \\ 
Chameleon       & 2,277           & 36,101          & 3                & 3,132   &     0.48       \\ 
Squirrel        & 5,201           & 217,073         & 3                & 31,48  &    0.42         \\ 
Email           & 1,005           & 25,571          & 42               & - &    0.40             \\ 


Cora            & 2,708           & 10,556          & 7                & 1433 &     0.24        \\ 
Citeseer        & 3,327           & 9,104           & 6                & 3703  &    0.14         \\ 
PPI             & 2,361           & 7,182           & 13               & -        &   0.13        \\ 
Actor     & 7,600 & 33,544  & 5       & 931      & 0.08           \\ 

\hline
\end{tabular}
\label{tab:dataset}
\end{table}
\textbf{Datasets.}
\label{sec:datasets}
We experimented with eight graphs from different domains such as actor co-occurrence network in films (Actor), airport networks (USAir), community network (Email), biological network (PPI), wikipedia networks (Chameleon and Squirrel), and citation networks (Cora and Citeseer). 
All these graphs have been used previously in different research papers \cite{kipf2016semi,musae,ribeiro2017struc2vec,pei2020geom,bo2021beyond}. 
Table \ref{tab:dataset} provides a summary of the graphs. The clustering coefficient of a node in a network measures the propensity to form a cluster and the Average Clustering Coefficient (ACC) is a generalization that represents this score for the whole network \cite{saramaki2007generalizations}. 
For networks without input features, we use a one-hot encoding representation of vertices and feed it to the GNNs as input features. 

{\bf Experiment settings.} 
Among the graphs having input features, Cora and Citeseer have high assortativity whereas Chameleon and Squirrel have low assortativity \cite{bo2021beyond}. Thus, for the classification task using Cora and Citeseer networks, we use 30 vertices per class for training, 500 vertices for validation, and 1000 vertices for testing. For all other networks, we use 70\% vertices for training, 10\% vertices for validation, and the remaining 20\% vertices for testing. 

In clustering experiments, we infer the predicted labels of all nodes using the trained models and compute clustering measures such as the Adjusted Rand Index. Similarly, for visualization, we infer 64-dimensional embeddings from the last convolutional layer of GNN models and generate 2D layouts using t-SNE~\cite{van2008visualizing}. Table \ref{tab:parametersforothers} reports a list of parameters used to run GNN methods.

\begin{figure*}[!htb]
\centering
     \includegraphics[width=0.23\linewidth,height=3.05cm]{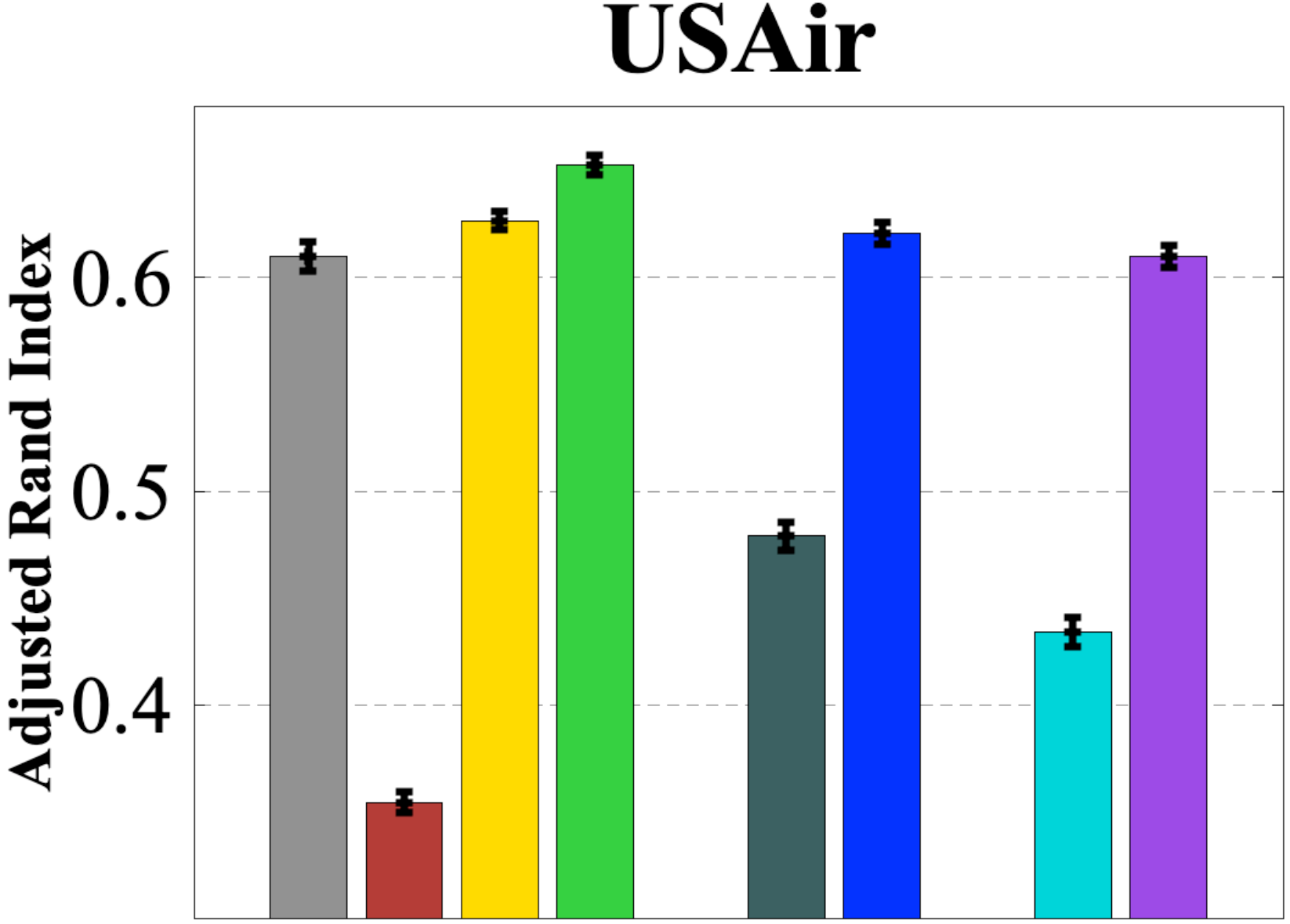}
     \includegraphics[width=0.23\linewidth,height=3.05cm]{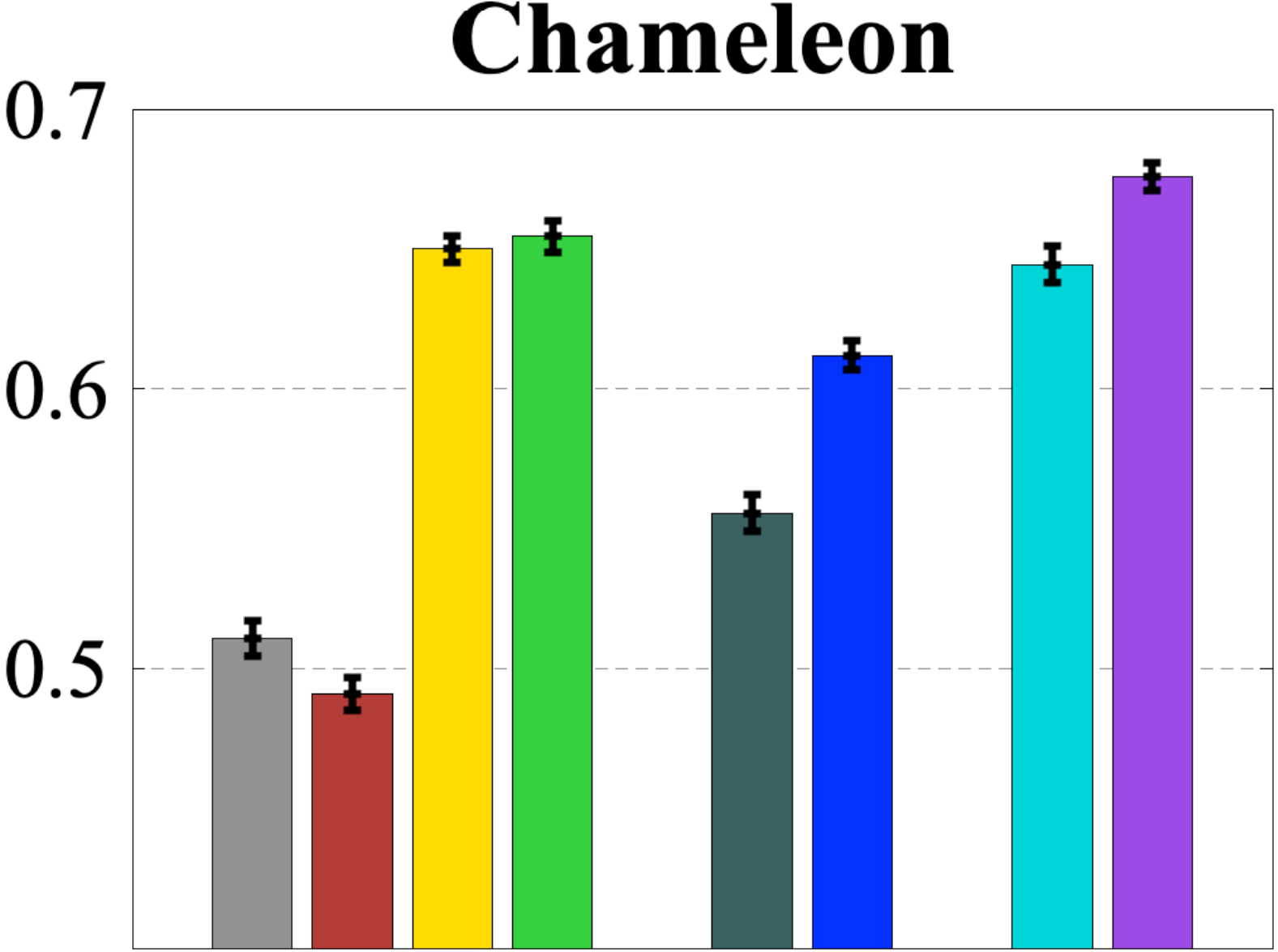}
     \includegraphics[width=0.23\linewidth,height=3.05cm]{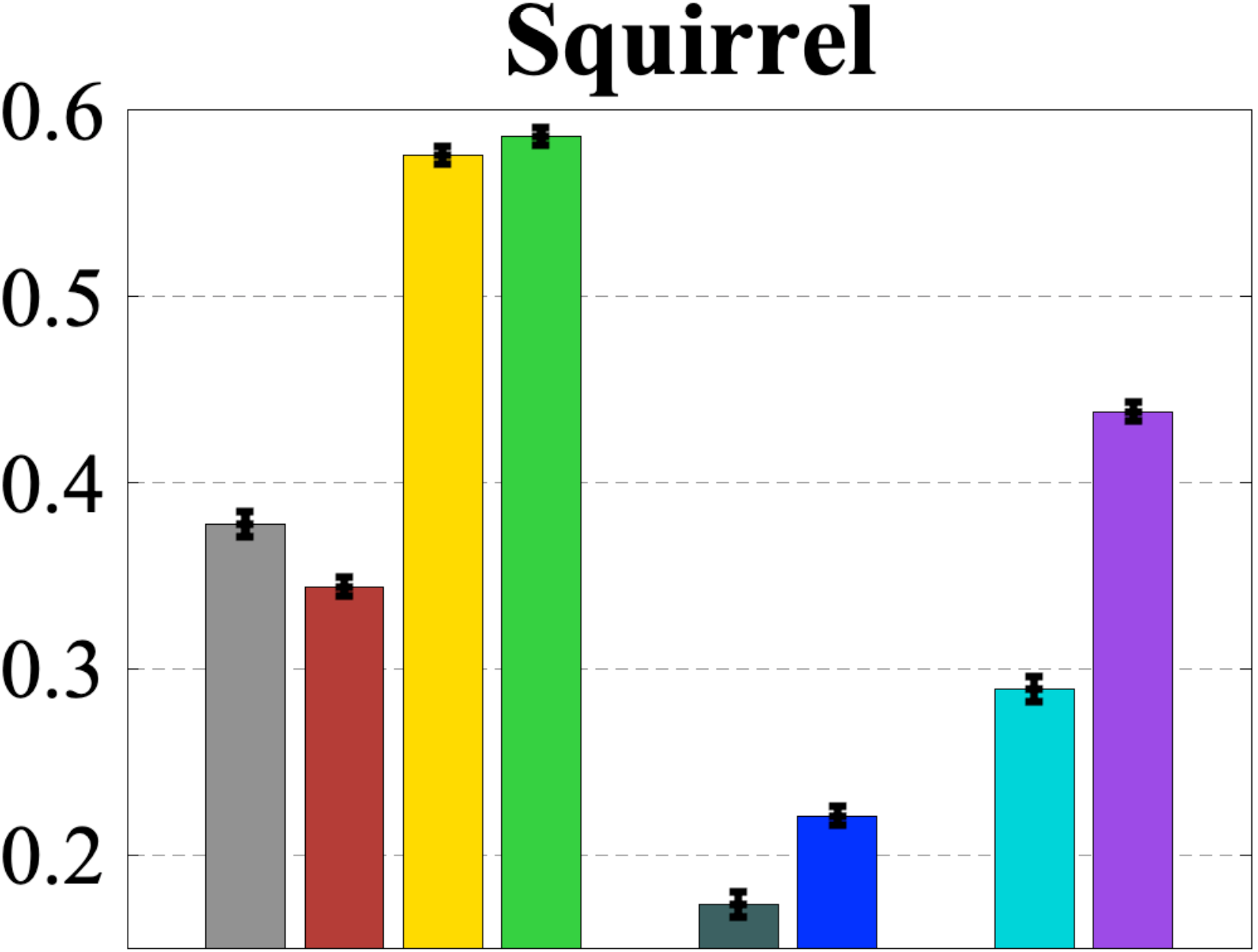}
      \includegraphics[width=0.23\linewidth,height=3.05cm]{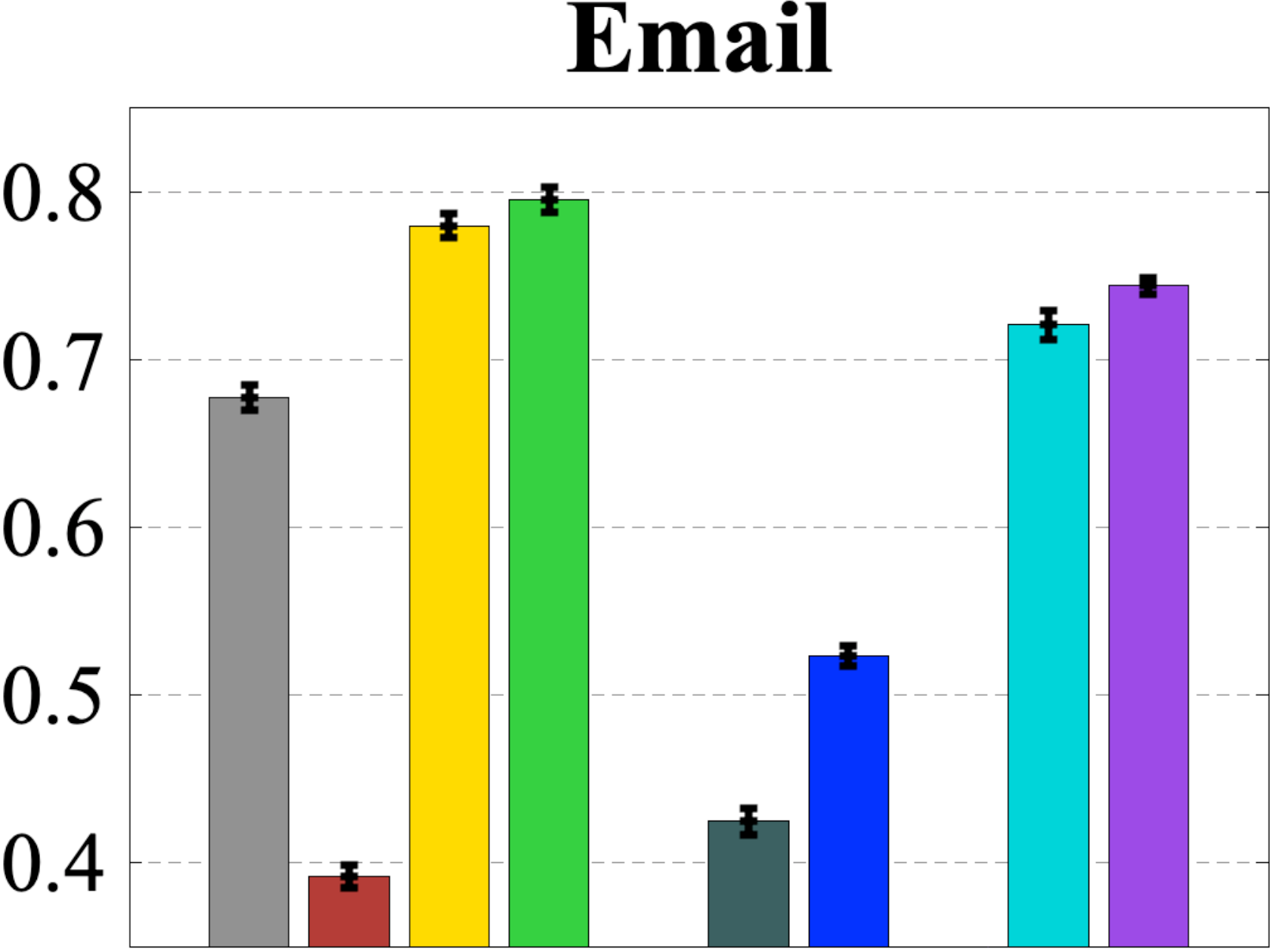}
       \includegraphics[width=0.23\linewidth,height=3.05cm]{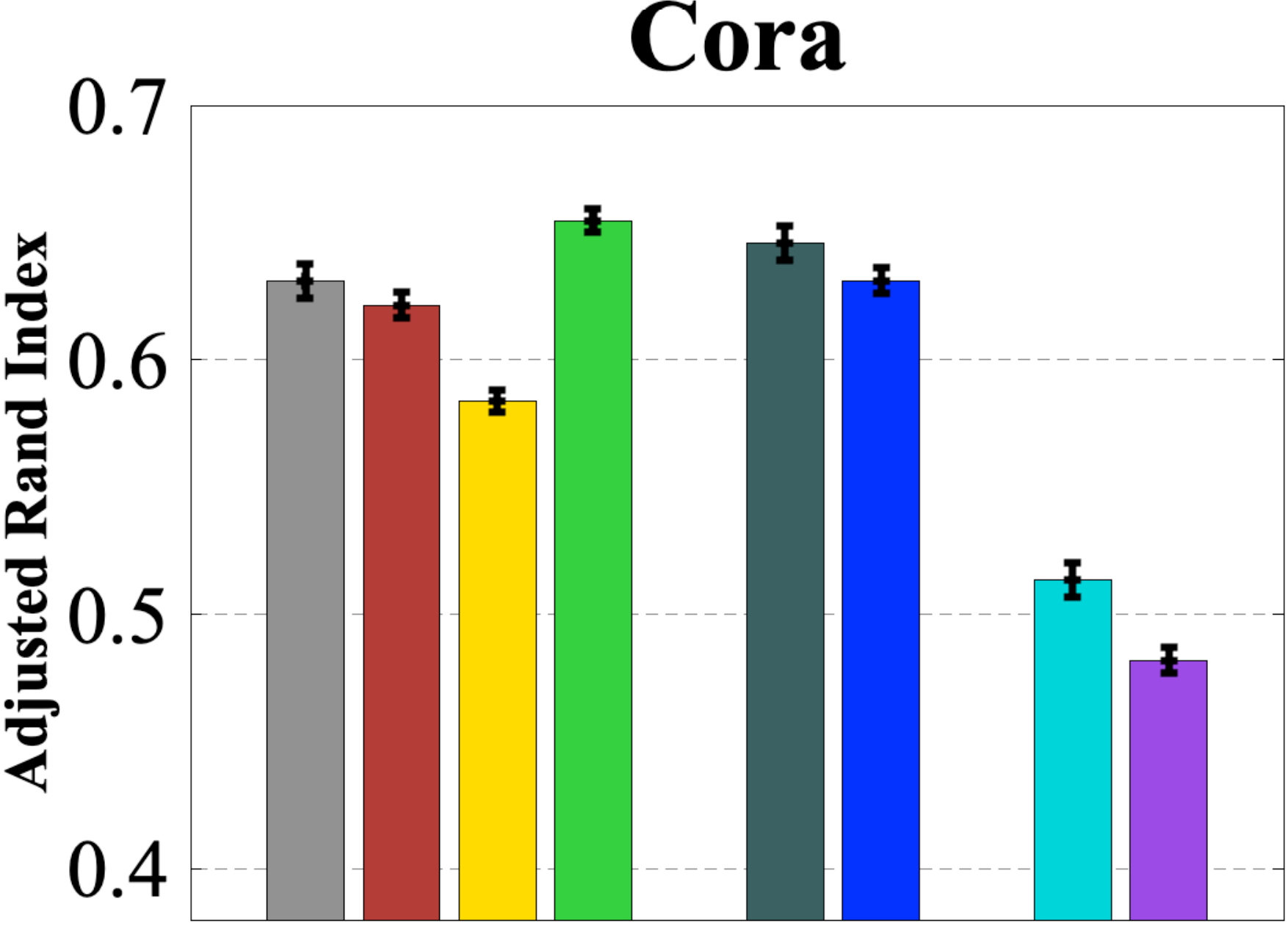}
        \includegraphics[width=0.23\linewidth,height=3.05cm]{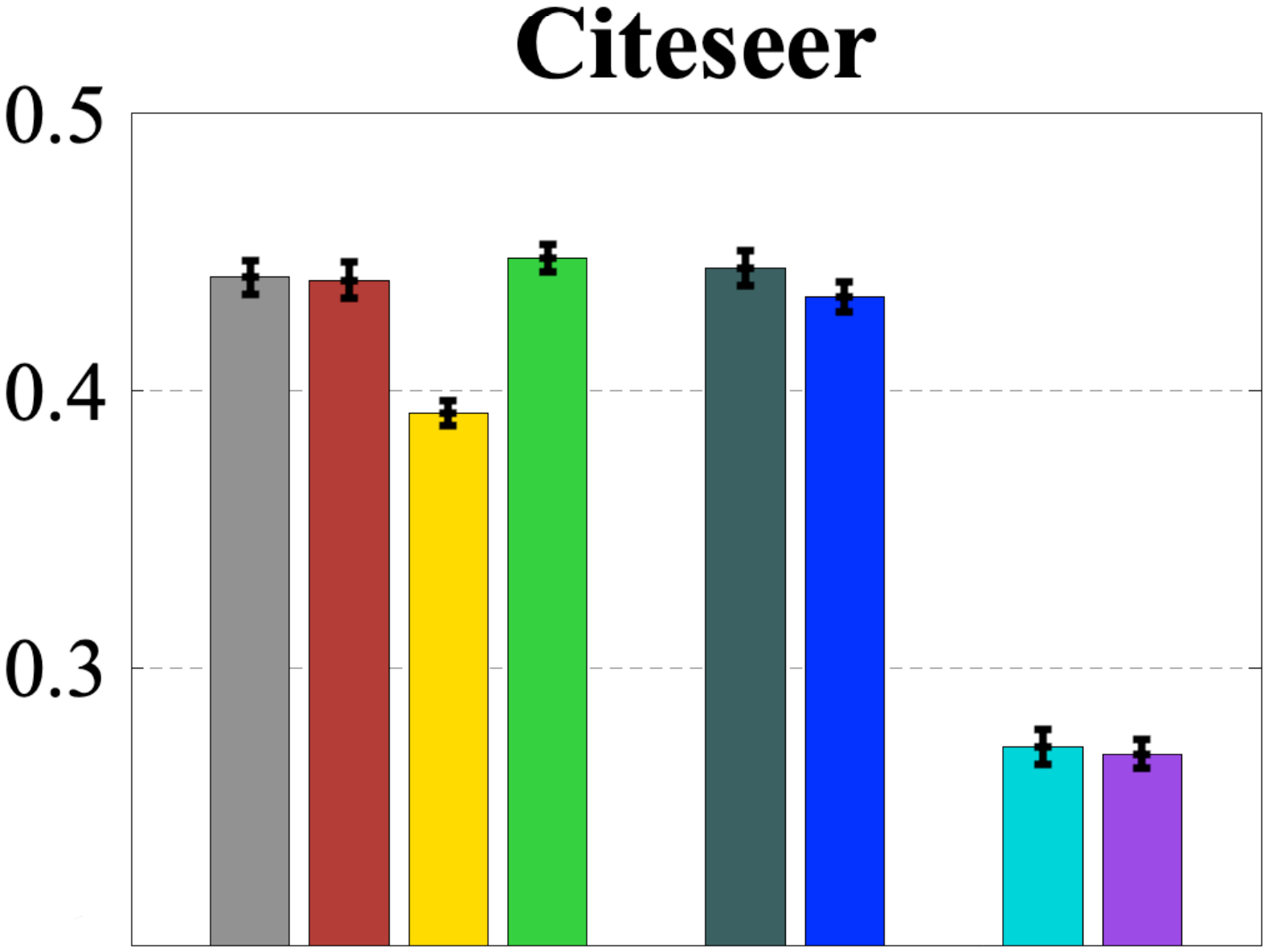}
    \includegraphics[width=0.23\linewidth,height=3.05cm]{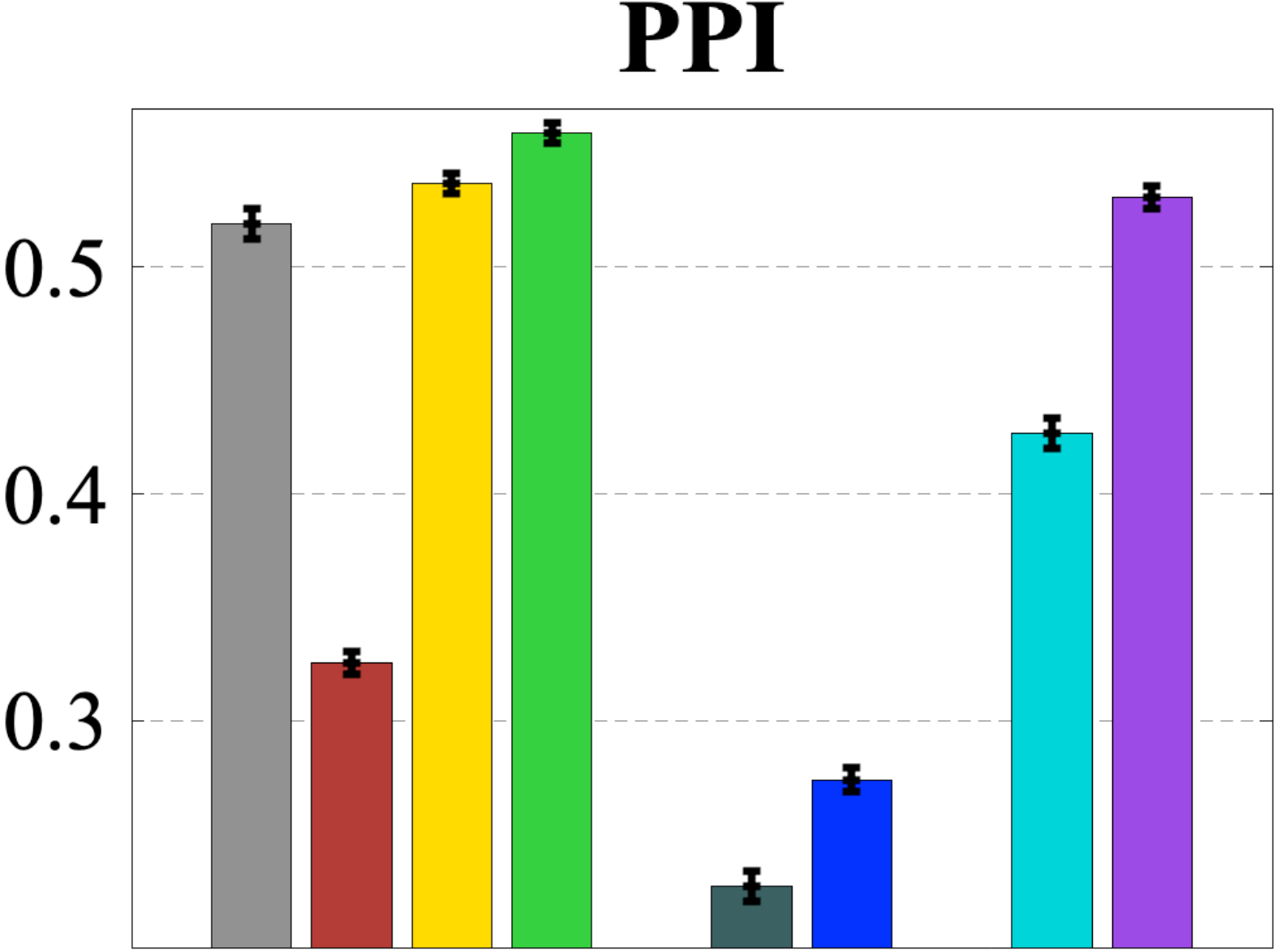}
     \includegraphics[width=0.23\linewidth,height=3.05cm]{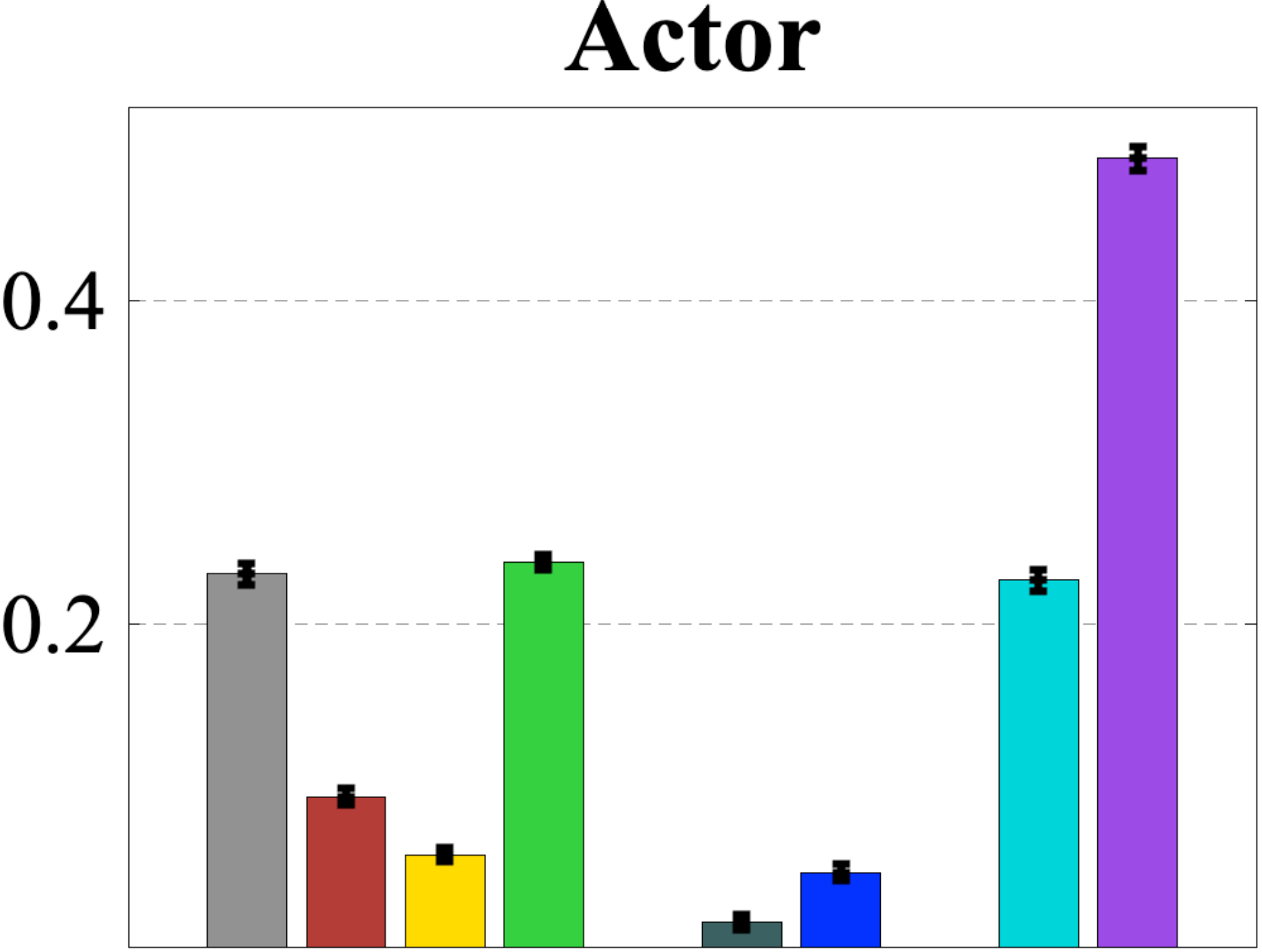} 
    \includegraphics[width=0.5\linewidth]{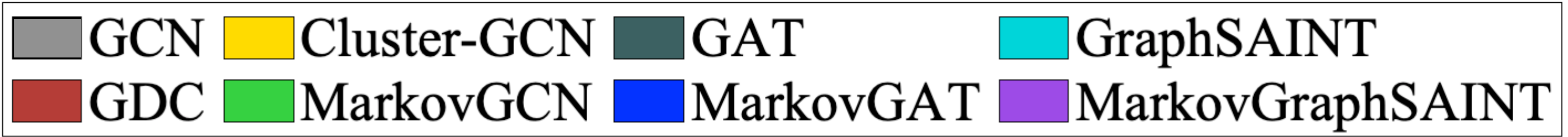}
    \caption{Comparing the quality of predicted clustering using the ARI index (higher is better). In each plot, bars are organized in three groups based on their base GNN models: GCN (first four bars), GAT (5th and 6th bars), and GraphSAINT (the last two bars). MarkovGNN represents the last bar in each group, and it generally improves the performance of the base GNN model.}
    \label{fig:ariresults}
\end{figure*}

\textbf{Baselines.} We choose five semi-supervised methods (GCN, GDC, GAT, Cluster-GCN, and GraphSAINT) to compare the effectiveness of proposed method. 
GCN is the pioneering work using graph convolutions in GNN models \cite{kipf2016semi}. 
GAT uses an attention mechanism by considering the influence of neighbors of a vertex in a graph \cite{velivckovic2017graph}. GDC uses diffusion on a graph (as shown in Eq.~\ref{eq:gdc}) and then feeds the modified graph to GNNs \cite{klicpera2019diffusion}. 
In our analysis, we consider GCN as an underlying method for GDC. Cluster-GCN partitions the graph using a clustering technique before feeding it to GNN \cite{chiang2019cluster}. 
GraphSAINT is a recent method based on different types of sampling approaches such as random-walk sampling and edge sampling \cite{zeng2019graphsaint}.
Besides these semi-supervised methods, we also experimented with two unsupervised graph embedding methods such as DeepWalk \cite{perozzi2014deepwalk} and struc2vec \cite{ribeiro2017struc2vec}. 
We set different hyper-parameters of other GNN models in PyG framework. Tables~\ref{tab:parametersforothers} and \ref{tab:markovgcnparams} provide a summary of the parameters used in our experiments.

\vspace{-0.35cm}
\subsection{Clustering Performance of MarkovGNN}
Theoretically, we expect that MarkovGNN would detect the clustering pattern present in the original graphs.
To validate this expectation, we use predicted vertex labels from MarkovGNN and other GNN methods to partition the graph into predicted clusters. 
We then use true class labels as ground truth clusters and measure the similarity between predicted and ground-truth clustering patterns.

{\bf Cluster similarity metrics.} 
To measure the similarity between the predicted and given clusterings, we use the adjusted Rand index (ARI) \cite{hubert1985comparing} and V-Measure \cite{rosenberg2007v}.
The original Rand index identifies (1) the number of vertex pairs that are in the same cluster in both given and predicted clusterings (true positives) and  (2) the number of vertex pairs that are in different clusters in given and predicted clusterings (true negatives).
The true positives and true negatives are added and normalized by the total number of vertex pairs to get the Rand index. 
ARI simply adjusts the Rand index for the chance grouping of elements.
The higher the ARI the more similar two clusterings are. The V-Measure~\cite{rosenberg2007v} is computed from the homogeneity ($h$) and completeness ($c$) scores. Homogeneity computes how different the predicted labels are for each true cluster and completeness computes how different the clusters are for each type of
predicted label. 
The V-Measure computes the weighted harmonic mean of $h$ and $c$ i.e., $\frac{(1+\beta)hc}{\beta h + c}$, where $\beta {>} 1$ puts more weight on homogeneity and $\beta {<} 1$ puts more weight on completeness (in our case, $\beta {=} 1$). A higher value of V-Measure represents more similar clusterings.

\begin{figure*}[!thb]
\centering
    \includegraphics[width=0.23\linewidth,height=3.2cm]{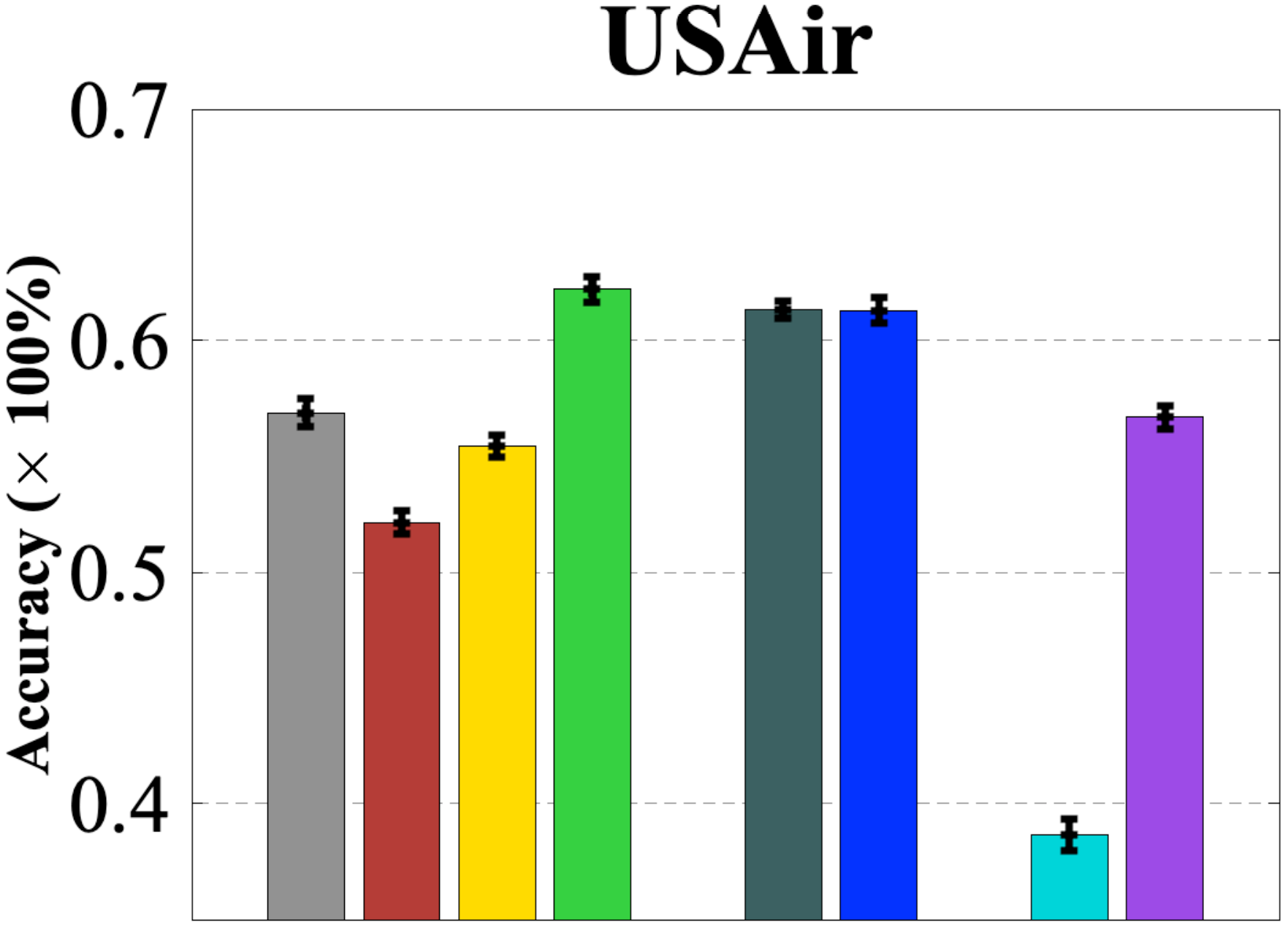}
    \includegraphics[width=0.23\linewidth,height=3.2cm]{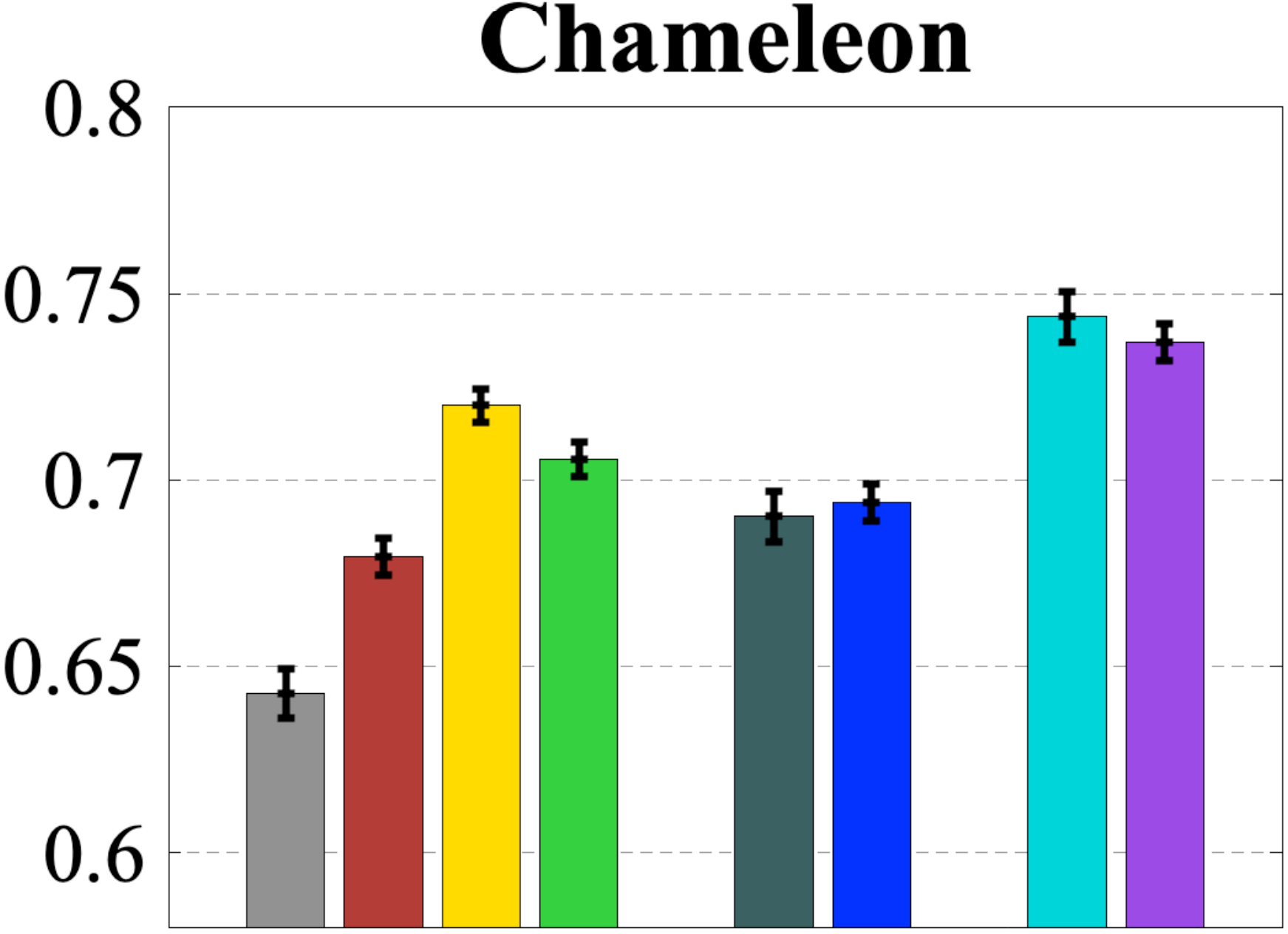}
    \includegraphics[width=0.23\linewidth,height=3.2cm]{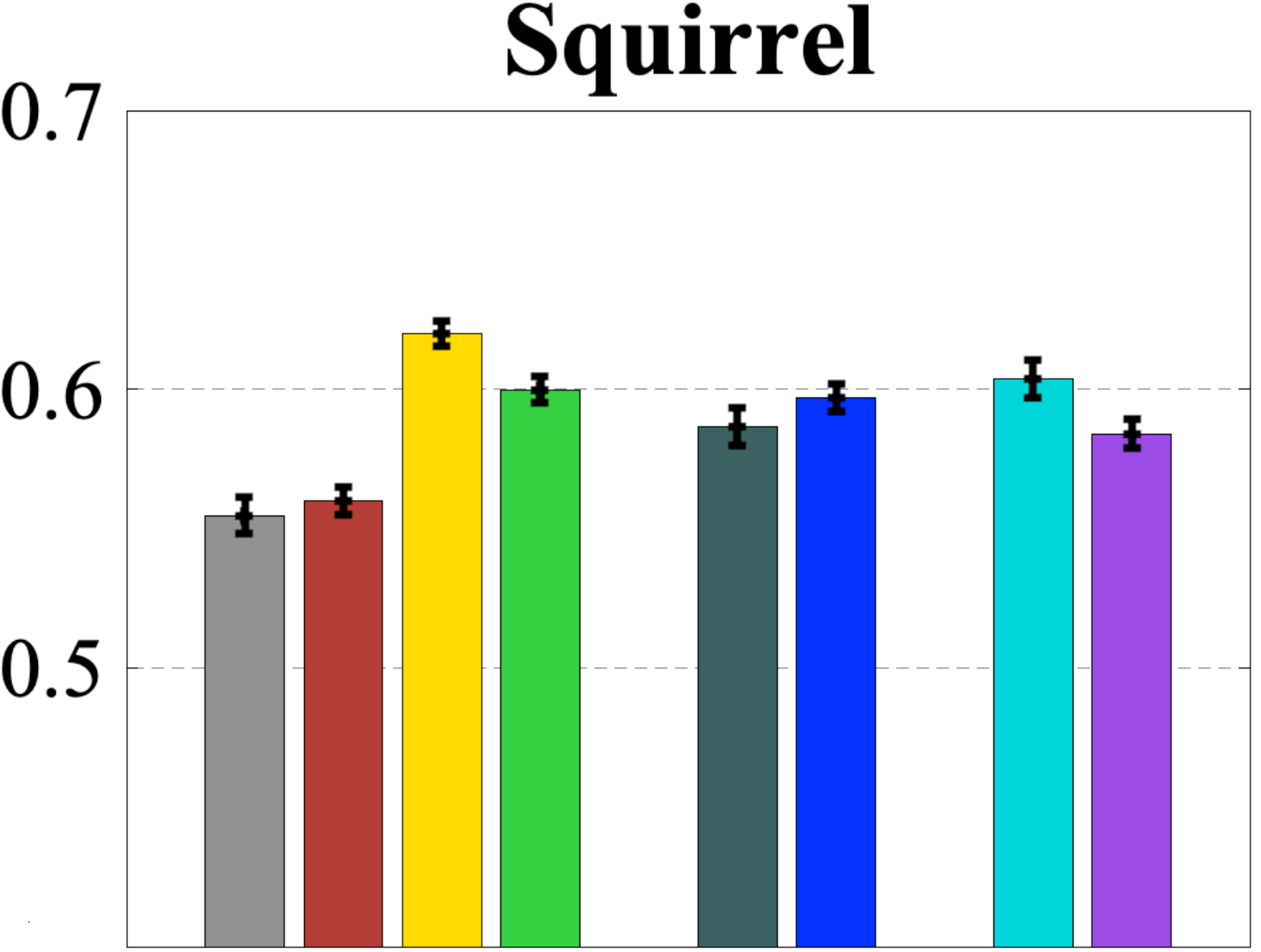}
    \includegraphics[width=0.23\linewidth,height=3.2cm]{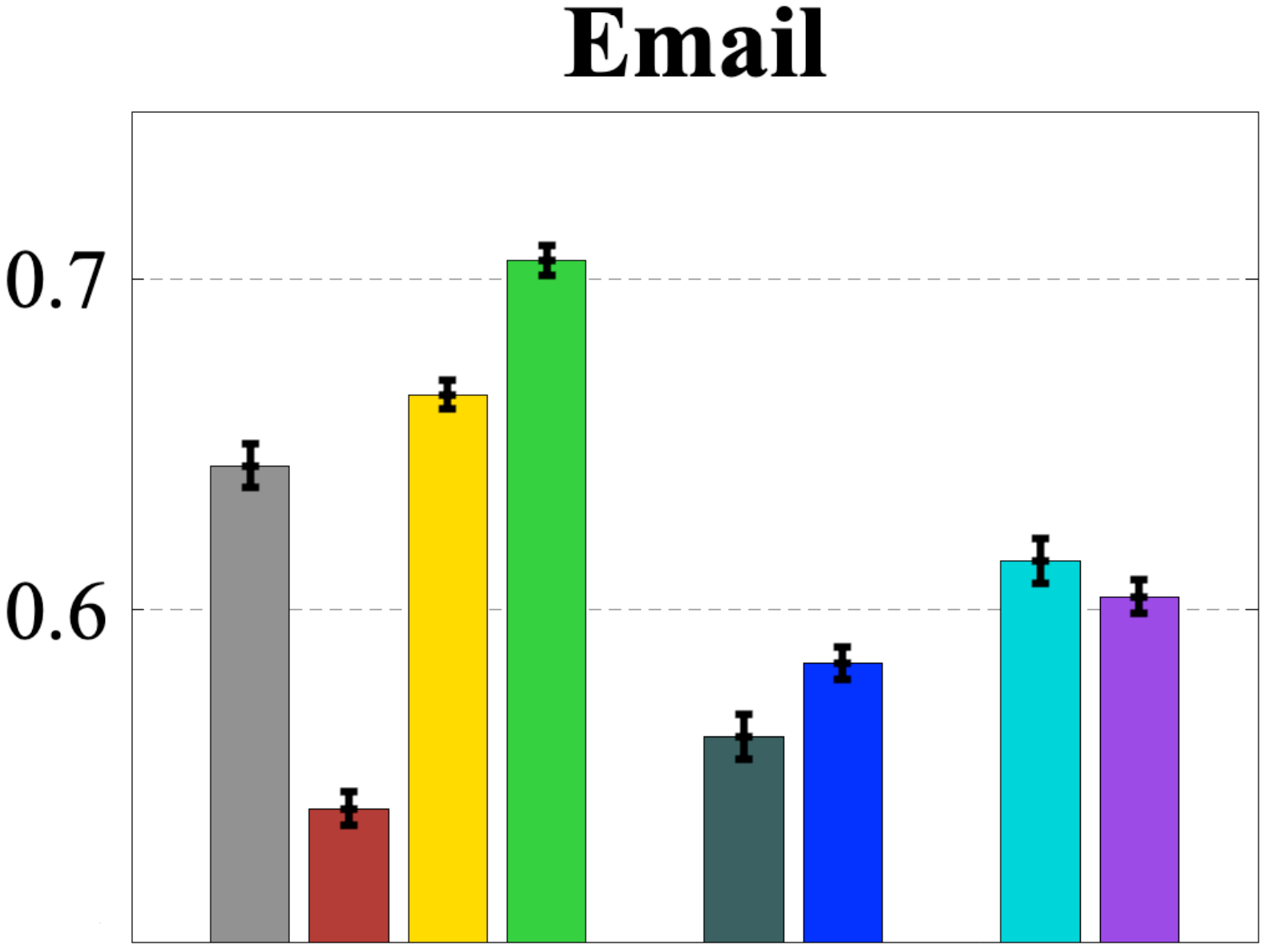}
    \includegraphics[width=0.23\linewidth,height=3.2cm]{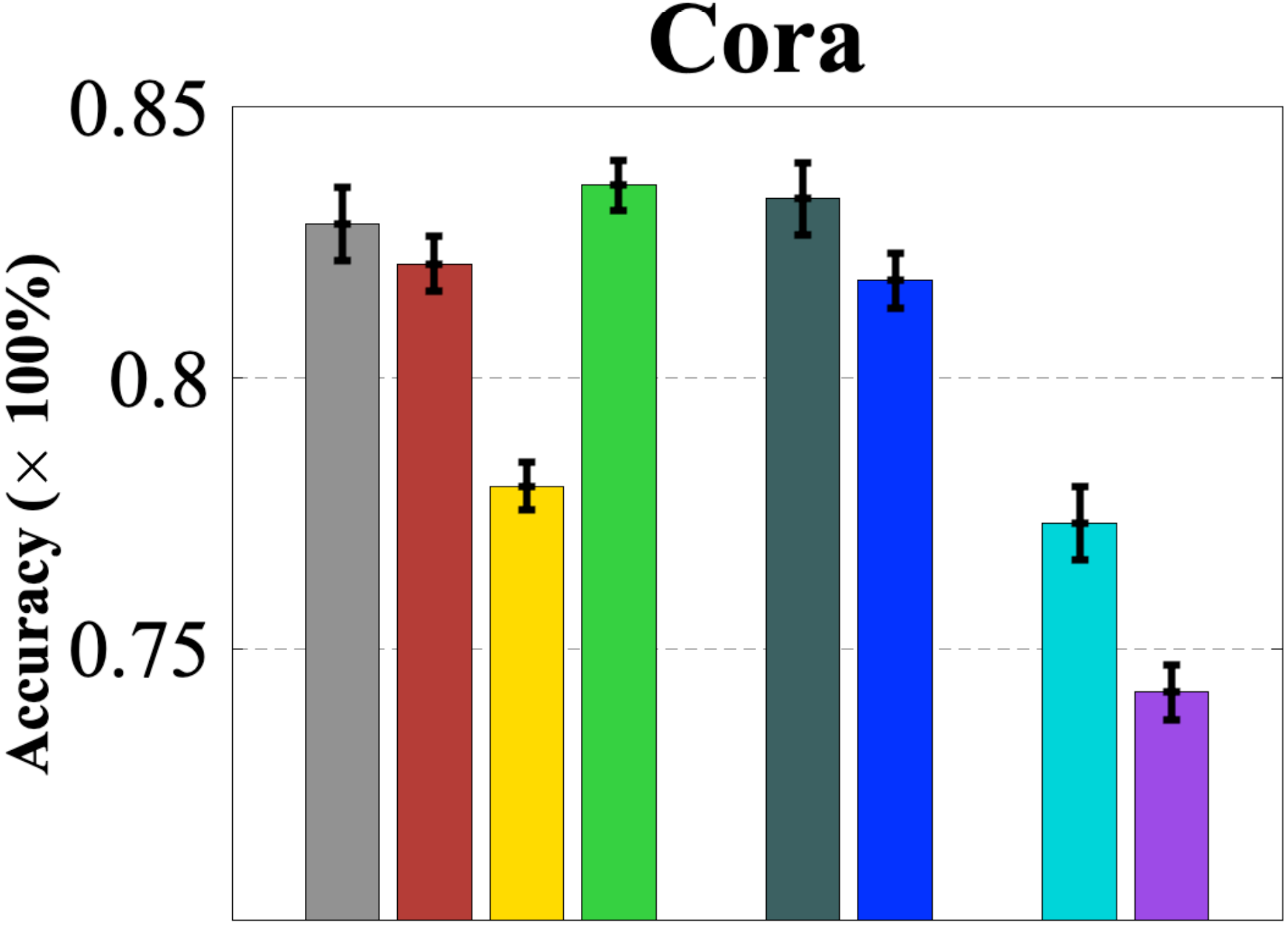}
     \includegraphics[width=0.23\linewidth,height=3.2cm]{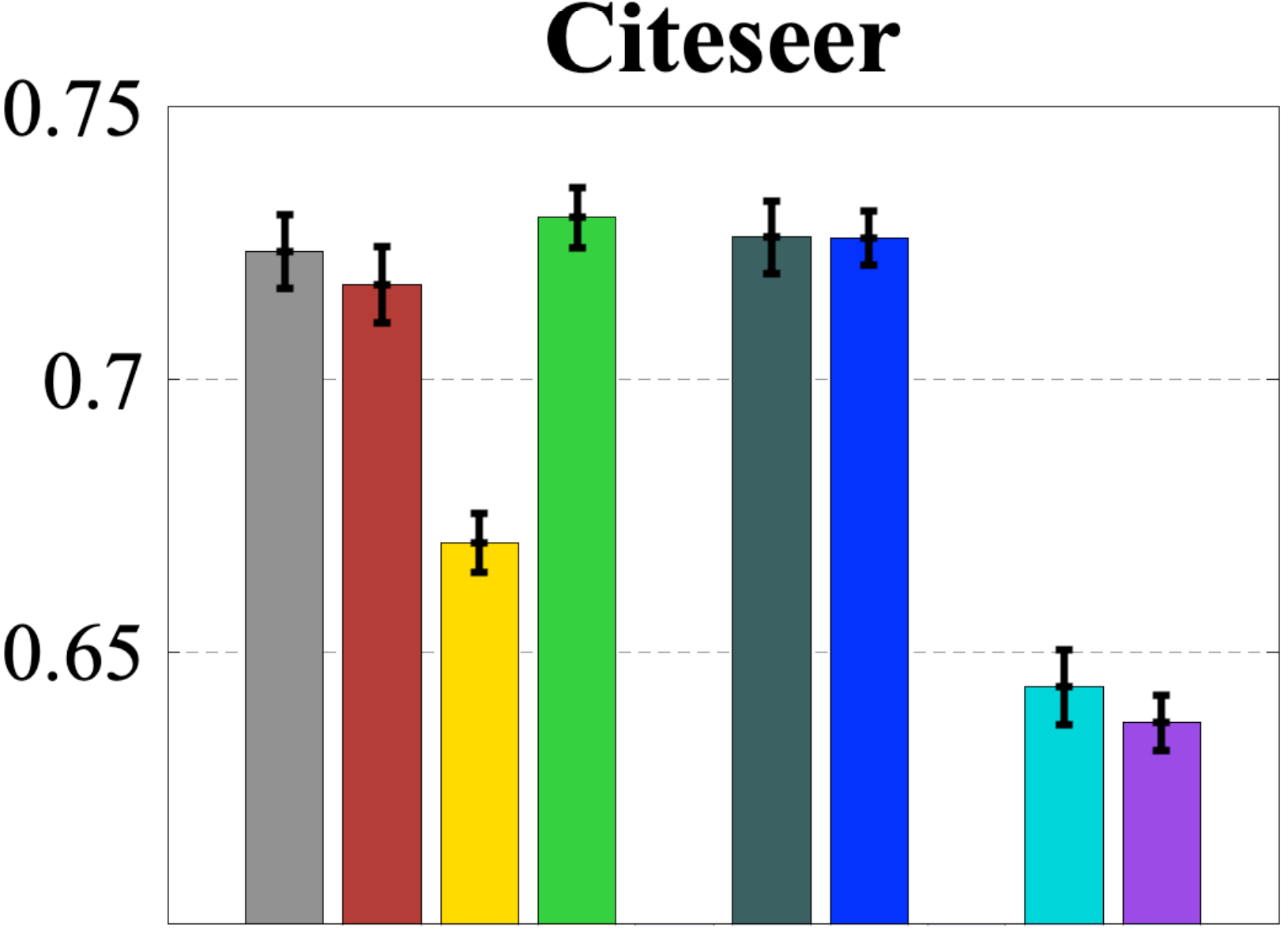}
      \includegraphics[width=0.23\linewidth,height=3.2cm]{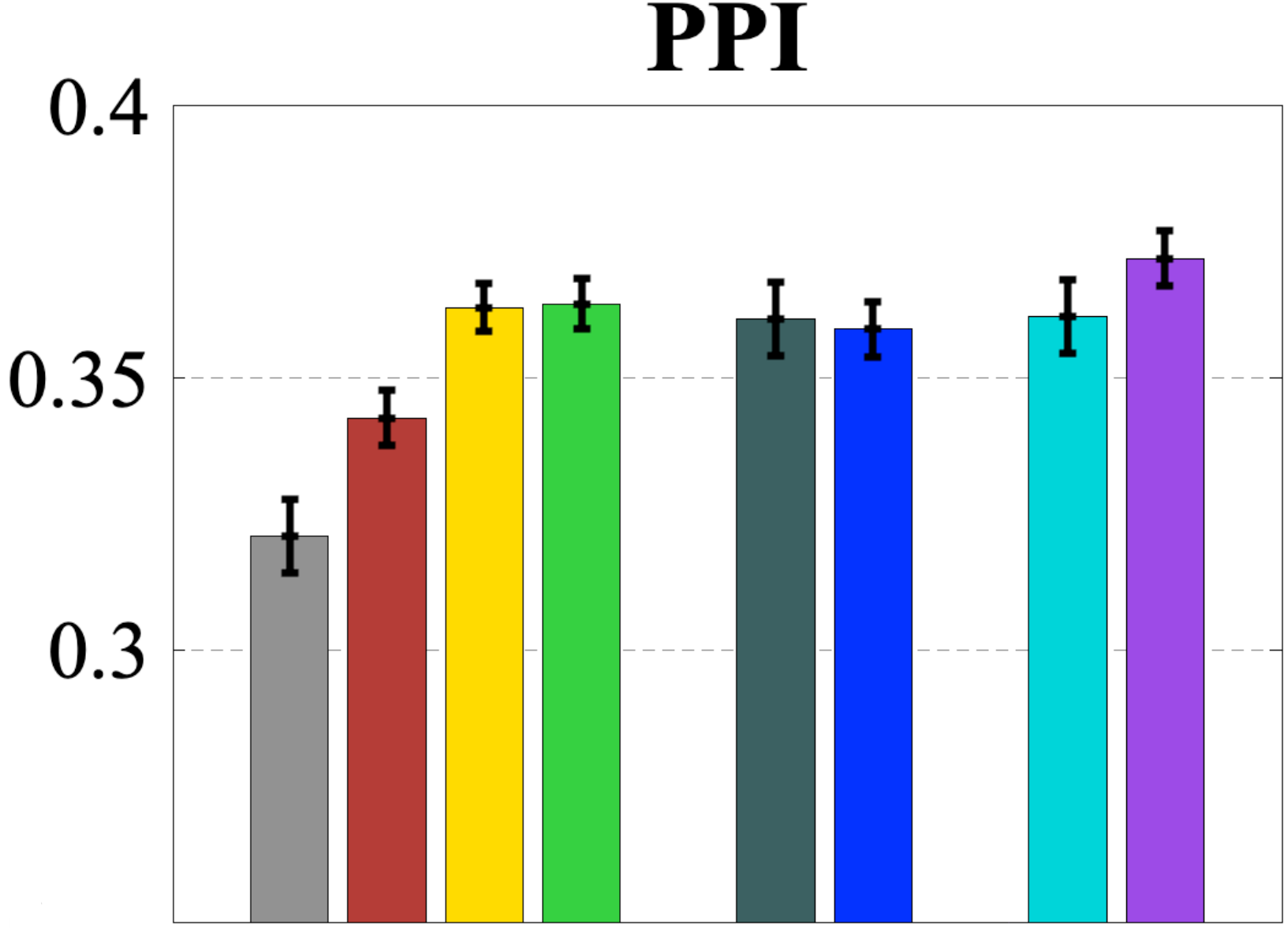}
    \includegraphics[width=0.23\linewidth,height=3.2cm]{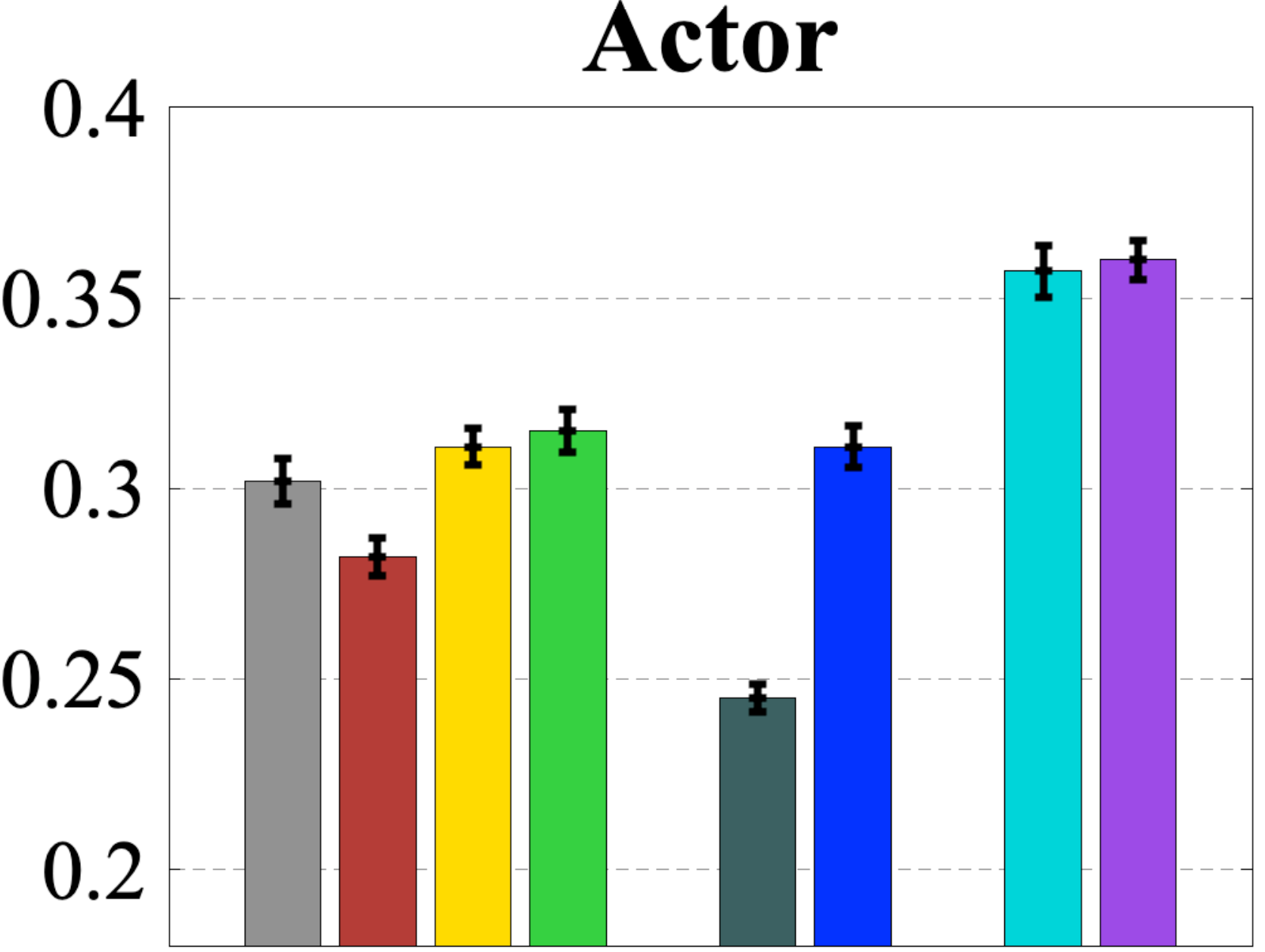} 
    \includegraphics[width=0.5\linewidth]{figures/legends2.pdf}
    \caption{Comparing the performance in classifying nodes using different variants of MarkovGNN and other GNN methods. In each plot, bars are organized in three groups based on their base GNN models: GCN (first four bars), GAT (5th and 6th bars), and GraphSAINT (the last two bars). MarkovGNN represents the last bar in each group, and it often improves the performance of the base GNN model.}
    \label{fig:accuracyresults}
\end{figure*}
{\bf MarkovGNN improves the prediction of clusters.}
Fig.~\ref{fig:ariresults} compares the quality of clustering predictions (using ARI) from various GNN methods. 
We observe that the Markov process improves the clustering performance of GCN, GAT, and GraphSAINT for almost all graphs. 
For example, MarkovGCN performs better than GCN for all graphs, MarkovGAT performs better than GAT for all graphs except Cora and Citeseer, and MarkovGraphSAINT performs better than GraphSAINT for all graphs except Cora and Citeseer.
Furthermore, MarkovGCN consistently outperforms GDC and Cluster-GCN even though the latter two methods use diffusion and clustering to preprocess the graph. 
Overall, for any graph, there is at least one variant of MarkovGNN that performs better than any other approach according to the ARI measure. 
For example, MarkovGCN performs the best for Cora even though Markov matrices do not improve the performance of GraphSAINT and GAT on this graph. 
MarkovGNN also exhibits better clustering performance according to V-Measure (expect for the PPI network where GraphSAINT performs the best) as shown in Table \ref{tab:clustering}.
We also experimented with unsupervised embedding methods DeepWalk, and struc2vec for community detection (see Fig. \ref{tab:deepwalk_struc2vec}). 
However, the unsupervised methods perform much worse than GNNs, 
which is not surprising because unsupervised methods do not use class labels when finding the embedding or communities.


\begin{table}[!htb]
\centering
\vspace{-0.1cm}
\caption{V-Measure (avg. of 10 runs) for different methods. A higher value of V-Measure means a better prediction of clusters. The best value for each dataset is highlighted in green. 
Clust.GCN - Cluster-GCN.
}
\vspace{-0.25cm}
\label{tab:clustering}
\begin{tabular}{p{1.65cm}|p{0.458cm}p{0.458cm}p{0.458cm}p{0.458cm}p{0.458cm}p{0.458cm}p{0.458cm}p{0.458cm}p{0.468cm}} 
\hline
Methods     & USAir                                     & Cham.                                     & Squi.                                     & Email                                    & Cora                                      & Cite.                                     & PPI                                       & Actor                                      \\ 
\hline
GCN         & 0.564                                     & 0.436                                     & 0.297                                     & 0.778                                    & 0.616                                     & 0.418                                     & 0.534                                     & 0.211                                      \\ 

GAT         & 0.443                                     & 0.479                                     & 0.152                                     & 0.653                                    & 0.636                                     & 0.425                                     & 0.318                                     & 0.014                                      \\ 

Clust.GCN & 0.591                                     & 0.532                                     & 0.38                                      & {\cellcolor[rgb]{0.851,0.918,0.827}}0.86 & 0.588                                     & 0.383                                     & 0.609                                     & 0.058                                      \\ 

GDC         & 0.263                                     & 0.414                                     & 0.26                                      & 0.634                                    & 0.629                                     & 0.418                                     & 0.359                                     & 0.091                                      \\ 

GraphSAINT  & 0.448                                     & 0.566                                     & 0.217                                     & 0.817                                    & 0.517                                     & 0.315                                     & {\cellcolor[rgb]{0.851,0.918,0.827}}0.673 & 0.221                                      \\ 

MarkovGCN   & {\cellcolor[rgb]{0.851,0.918,0.827}}0.607 & {\cellcolor[rgb]{0.851,0.918,0.827}}0.576 & {\cellcolor[rgb]{0.851,0.918,0.827}}0.493 & {\cellcolor[rgb]{0.851,0.918,0.827}}0.86 & {\cellcolor[rgb]{0.851,0.918,0.827}}0.638 & {\cellcolor[rgb]{0.851,0.918,0.827}}0.426 & 0.601                                      & {\cellcolor[rgb]{0.851,0.918,0.827}}0.228  \\
\hline
\end{tabular}
\arrayrulecolor{black}
\end{table}
{\bf When does MarkovGNN excel in predicting clusters?}
When a graph has a high clustering coefficient, MarkovGNN predicts the original clustering much better than its peers.
This behavior can be observed in Fig.~\ref{fig:ariresults} and Table~\ref{tab:clustering}. 
For example, in Fig.~\ref{fig:ariresults}, MarkovGCN performs much better than GCN for the USAir, Chameleon, Squirrel, and Email graphs that have clustering coefficients greater than or equal to $0.4$ (see Table~\ref{tab:dataset}).
Cluster-GCN also performs well (slightly worse than MarkovGCN) for these graphs, indicating the usefulness of cluster-based GNN methods. 
However, for Cora, Citeseer, and PPI that have a clustering coefficient less than 0.25, MarkovGCN performs slightly better than GCN. 
Hence, the utility of MarkovGNN may diminish when the input graph lacks clustering patterns. 
MarkovGNN not only attains higher values for ARI and V-measure, it also achieves higher true positive rates for individual classes. See  Table~\ref{tab:chameleonconfusion} in the appendix for an example. 

\begin{figure}[!t]
    \centering
    \fbox{\includegraphics[width=0.46\linewidth,height=3.cm]{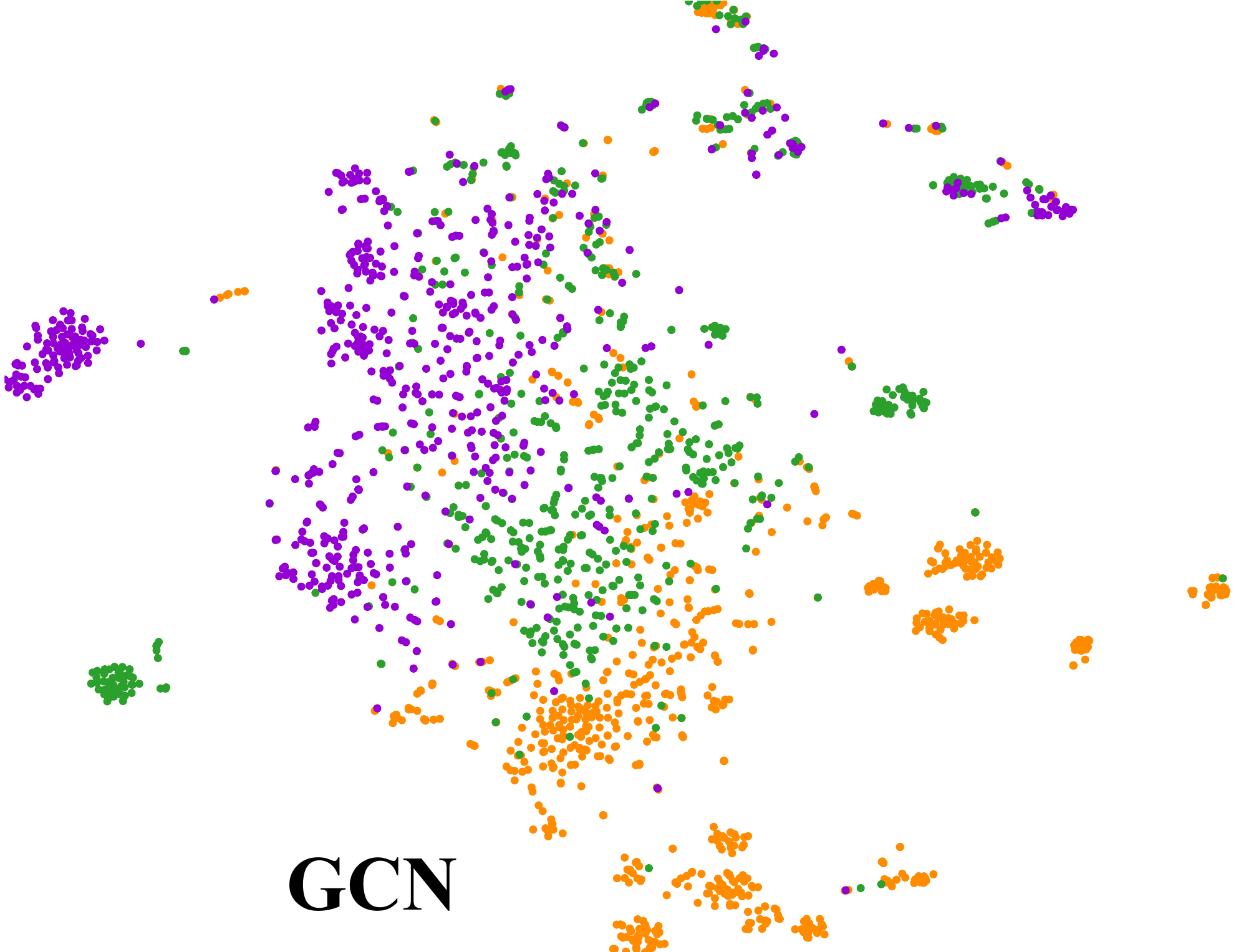}}
    \fbox{\includegraphics[width=0.46\linewidth,height=3.cm]{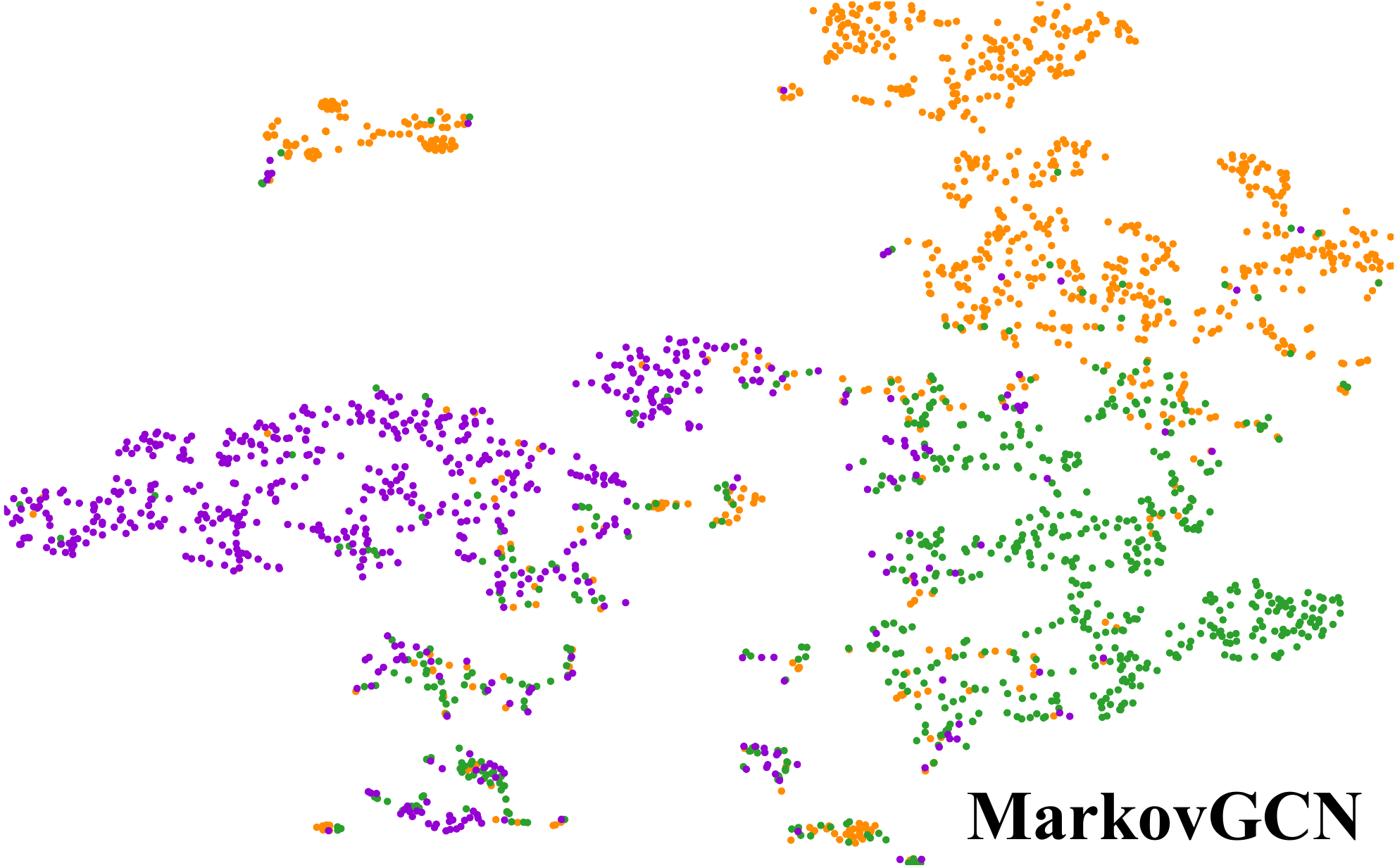}}
    \vspace{-0.35cm}
    \caption{2D layouts of the Chameleon graph from the embedding obtained from GCN in the left and MarkovGCN in the right.
    2D layouts are obtained using t-SNE. \vspace{-0.05cm}}
    \label{fig:visualization}
\end{figure}
\begin{figure*}[!t]
    \centering
    \fbox{\includegraphics[width=0.35\linewidth,height=4.cm]{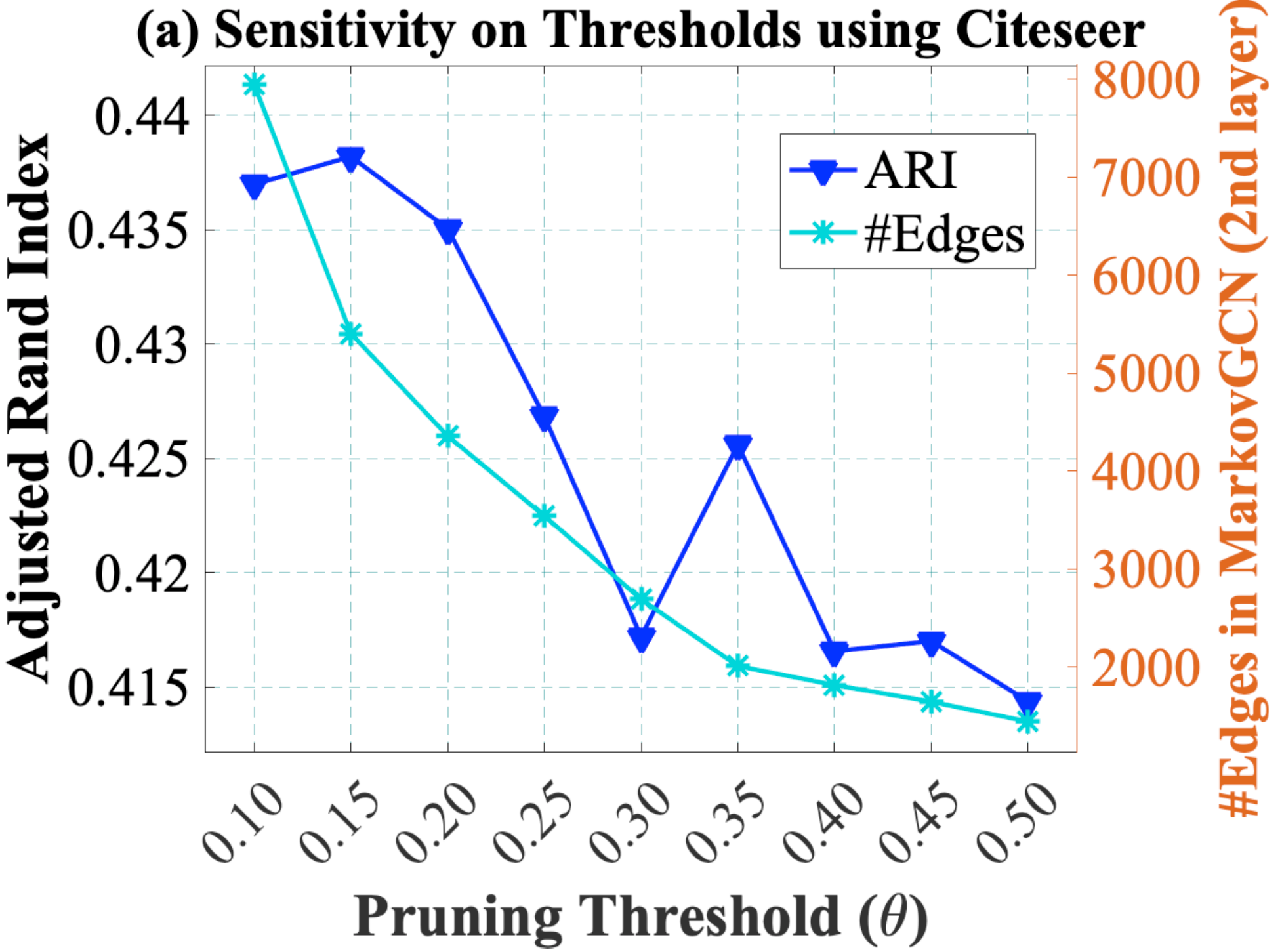}}
    \fbox{\includegraphics[width=0.26\linewidth,height=4.cm]{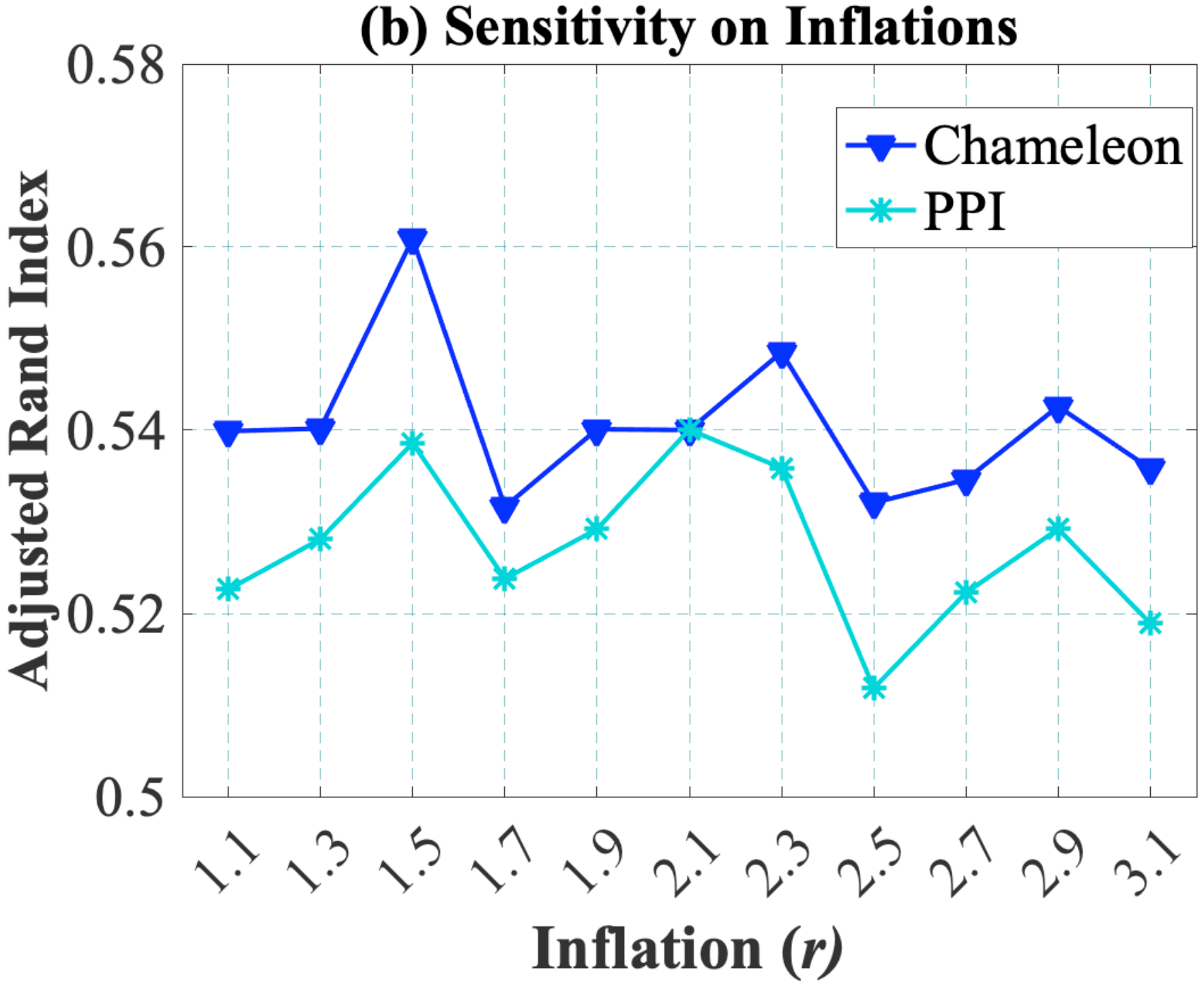}}
    \fbox{\includegraphics[width=0.29\linewidth,height=4.cm]{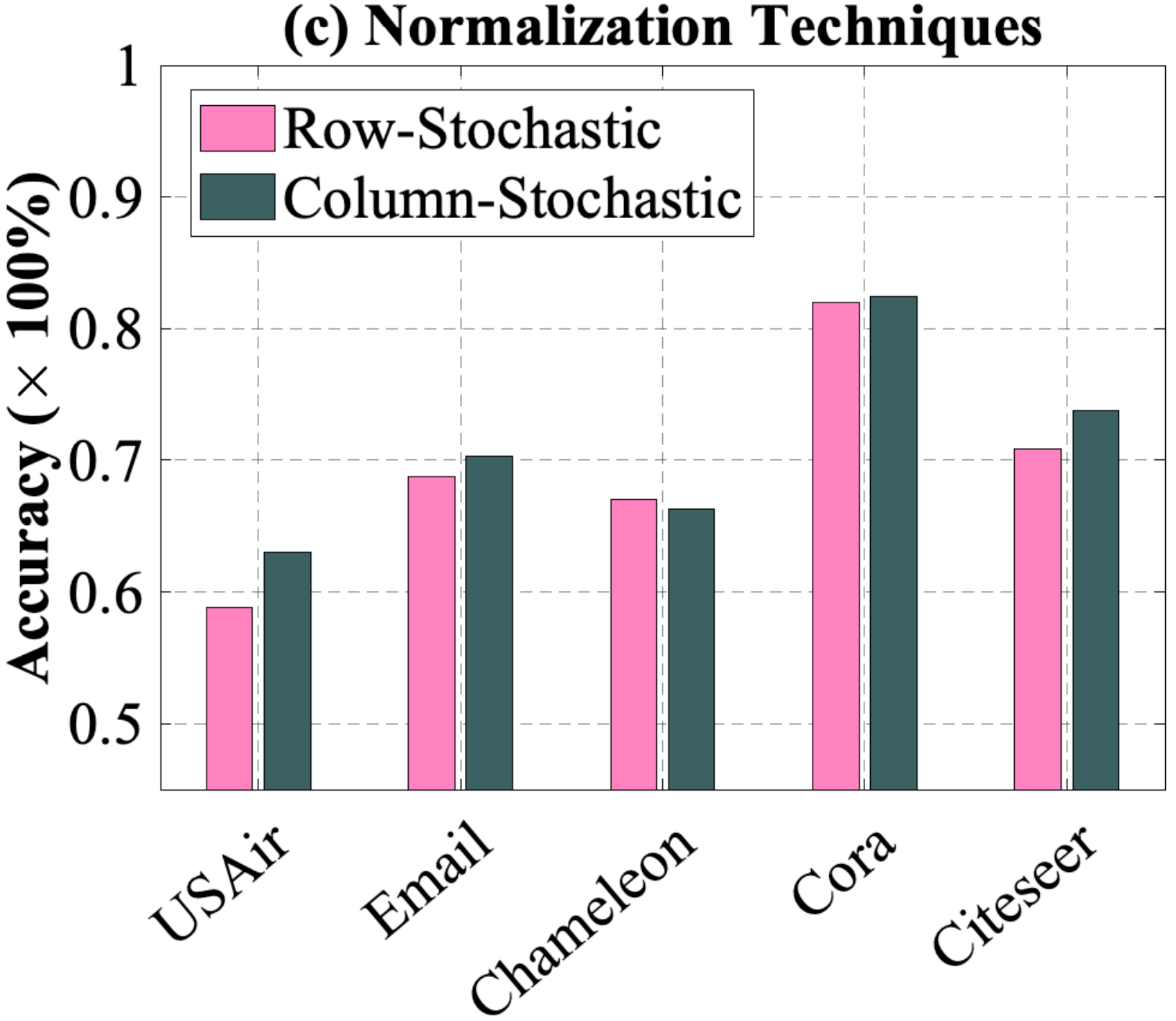}}
    \vspace{-5pt}
    \caption{Parameter sensitivity of MarkovGCN showing (a) influence of pruning threshold on Citeseer network with corresponding ARI and the remaining edges for the 2nd convolutional layer, (b) influence of inflation parameter for Chameleon and PPI networks, and (c) effect of using row vs. column stochastic matrices in Algorithm \ref{algo:mcl}.}
    \label{fig:paramsensitivity}
\end{figure*}
\subsection{Node Classification}
\vspace{-1pt}
Fig. \ref{fig:accuracyresults} shows the test accuracy in predicting node labels from different GNN methods.  
We observe that at least one variant of MarkovGNN shows the best test accuracy for all graphs except for Chameleon and Squirrel where GraphSAINT and Cluster-GCN, respectively, perform better than any other methods.
Generally, MarkovGNN improves the performance of its base GNN model in most graphs. 
In particular, MarkovGCN tends to classify nodes well for graphs where graph-structural properties significantly impact node classification such as USAir.  
Similarly, the Email network has known ground truth communities that  MarkovGCN can exploit to outperform other methods  significantly. 
For widely studied graphs such as Cora and Citeseer, MarkovGCN marginally outperforms GCN.
However, we do not expect significant improvement for Cora and Citeseer because of their lower average clustering coefficient values as shown in Table~\ref{tab:dataset}.
Note that GraphSAINT is a recent method that performs exceptionally well for Chameleon and Squirrel networks. 
However, GraphSAINT does not perform well (often worse than vanilla GCN) for graphs with known communities (e.g., the Email network) and structures (e.g., USAir).
By contrast, these are the networks where MarkovGCN excels.

\subsection{Visualization of clusters}
If an embedding of a graph captures the community structure, it should provide an aesthetically pleasing visualization by keeping nodes well clustered in the embedding space. 
To show the impact of MarkovGNN, we take the 64-dimensional embedding from the last layer of a GNN and use t-SNE~\cite{van2008visualizing} to project 64-dimensional embeddings to 2D space. 
We show the visualizations for the Chameleon graph from GCN and MarkovGCN in Fig.~\ref{fig:visualization}, where nodes from the same ground-truth class are shown with the same color.
Fig.~\ref{fig:visualization} shows that the visualization obtained from MarkovGCN  tends to place vertices from the same class together while keeping different classes well separated. 
For example, we can easily identify three clusters from the MarkovGCN-based visualization in the right subfigure, but the visualization from GCN failed to separate clusters from one another.  
\begin{table}[!htb]
\centering
\caption{Average runtimes of 10 runs in seconds are reported. For all GNNs, we only measure the training time of 200 epochs in the PyG framework. \vspace{-7pt}}

\begin{tabular}{c|p{0.65cm}p{0.65cm}p{0.65cm}p{1.27cm}|p{1.6cm}}
\hline
Graphs   & GCN    & GAT    & GDC    & Gr.SAINT & MarkovGCN \\ \hline
USAir    & 7.4   & 21.5  & 95.3  & 212.4     & 7.7       \\
Email    & 14.3  & 31.8  & 98.2  & 251.9     & 11.3      \\
PPI      & 13.7  & 32.6  & 228.6 & 167.2     & 17.3      \\
Chamel.  & 26.8  & 60.3  & 222.3 & 441.1     & 30.6      \\
Squirrel & 229.9 & 327.3 & 678.9 & 2,110.9   & 188.8     \\
Cora     & 10.9 & 31.2  & 245.8 & 160.8     & 16.4      \\
Citeseer & 24.1  & 67.9  & 317.8 & 220.3     & 41.7      \\ \hline
\end{tabular}
\label{tab:runtime}
\end{table}
\subsection{Running Time}

We measure the average training time of all methods for 10 runs (available in Table \ref{tab:runtime}).
Despite needing additional time in generating Markov matrices (Algorithm~\ref{algo:mcl}), MarkovGCN and GCN have comparable running time.  
On the other hand, GDC runs a diffusion step that is computationally similar to Markov clustering. 
However, effective sparsification by inflation and pruning makes MarkovGCN an order of magnitude faster than GDC.
Similarly, MarkovGCN runs $10\times$ to $100\times$ faster than GraphSAINT.
The advantage of MarkovGCN is originated from the decreasing number of edges in deeper layers.
By contrast, other GNN methods use the same graph in all layers. 

\subsection{Parameter Sensitivity}

\label{sec:paramsensitivity}

\textbf{The impact of the pruning threshold ($\theta$).} The pruning threshold plays an important role in the convergence of the Markov process. Fig. \ref{fig:paramsensitivity} (a) reports the ARI of MarkovGCN using Citeseer  for different values of $\theta$. 
For the Citeseer network, MarkovGCN achieves the best ARI score when  $\theta \approx 0.15$. 
We also observe in Fig. \ref{fig:paramsensitivity} (a) that an increasing threshold decreases the number of edges in the second layer of MarkovGCN.
Thus, the computation can be made much faster by sacrificing the accuracy slightly. 
If ARI or other measures do not change for different values of $\theta$, we prefer to use higher thresholds which typically make training and inference faster.






\textbf{Impact of the inflation parameter ($r$).} To find an optimal value of inflation in Algorithm \ref{algo:mcl}, we vary $r$ from 1.1 to 3.1 and run the experiments using Chameleon and PPI networks. Fig. \ref{fig:paramsensitivity}(b) shows that MarkovGCN is less sensitive to the inflation parameter. We observe better clustering results when $r$ is close to 1.5 and keep this value as a default value of $r$. 

\textbf{Column vs row stochastic matrices.} 
Algorithm~\ref{algo:mcl} maintains column stochastic matrices in every iteration. 
Fig. \ref{fig:paramsensitivity}(c)
shows that row stochastic matrices also  perform similarly for different graphs. 






\begin{figure}[!tb]
    \centering

  \includegraphics[width=0.6\linewidth]{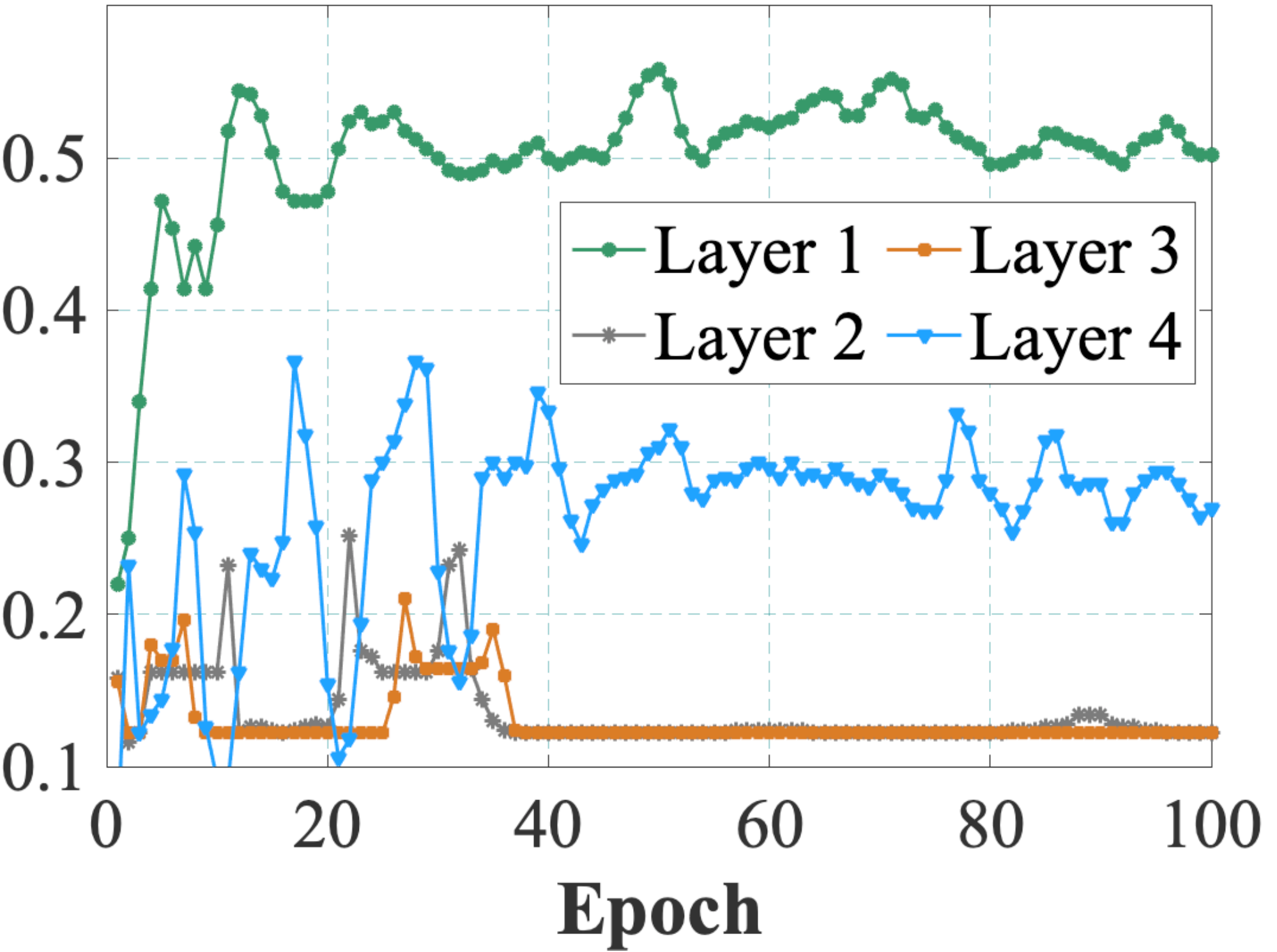}

    \caption{Layer-wise validation accuracy in GCN model for the Cora network.}
    \label{fig:initial_results}
\end{figure}

{\bf Residual connections with a hyper-parameter ($\alpha$).} 
In Eqn. \ref{eqn:residual}, we aim to control the residual connections with a hyper-parameter $\alpha$.
To select a suitable residual connection for $\phi$ in Eqn. \ref{eqn:residual}, we conduct an experiment to predict the impact of different layers on the predictive performance of a four-layer GNN. 
Following the protocol used by Raghu et al.~\cite{raghu2017expressive}, we disabled gradient updates for all layers except one and plot the observed validation accuracy for different layers in Fig. \ref{fig:initial_results}.
For the Cora network, we observe that the input layer (layer 1) has the highest accuracy followed by layer 4. 
The contribution of layer 1 is expected to be significant for graphs such as Cora that include node features.
Thus, when adding residual connections, we select layer 1 in Eqn. \ref{eqn:residual}. 

%% file: supplementary.tex
\newpage
\section{Experimental Details}
\label{sec:experimentaldetails}
\textbf{Software Versions.} We use Python programming language v3.7.3 to implement and perform all of our experiments. We use several python packages for our implementation which are outlined as follows: PyTorch: 1.7.0, PyTorch Geometric \cite{fey2019fast} version 1.6.1 along with PyTorch Sparse 0.6.8, PyTorch Scatter 2.0.5, PyTorch Cluster 1.5.8,  PyTorch Spline Conv 1.2.0, NetworkX: 2.2, scikit-learn: 0.23.2, Matplotlib: 3.0.3, python-louvain: 0.13

\textbf{Implementation Details.}
The proposed MarkovGCN model has been implemented in PyG framework\footnote{https://github.com/rusty1s/pytorch\_geometric}. We employ both $torch.mm$ and $torch\_sparse.spspmm$ functions to perform dense and sparse matrix multiplication operation, respectively, on line 4 of Algorithm \ref{algo:mcl}. We observe very little advantage of using the sparse matrix. Once MCL steps are done, we select $L$ number of matrices from $k$ matrices based on the type of MarkovGNN variant. Each convolutional layer of MarkovGNN has a different graph structure which is different from GCN. Otherwise, similar types of operations are performed in both forward and backward propagation as GCN using $GCNConv$ class. In $l$-th convolutional layer, we pass all edges of $A^{(l)}$ in MarkovGCN whereas GCN uses edges of the original matrix $A$. Similar to other GNN models, we employ the negative log-likelihood loss function and Adam optimizer of PyTorch package. Instead of the threshold-based pruning strategy in Algorithm \ref{algo:mcl}, we have also implemented a sparse $k$-selection technique where we keep top-$k$ weighted entries in each row of the stochastic matrix after each iteration of MCL. However, we have found from empirical results that this method is not very effective and thus we do not report here. To conduct the experiment for layer-wise contribution of Fig. \ref{fig:initial_results} (b), we set the $\mathbf{weight.requires\_grad}$ and $\mathbf{bias.requires\_grad}$ parameters of the corresponding layer to \emph{True} while setting other model parameters to \emph{False}.

\textbf{More Details on Dataset.}
We have collected USAir network from struc2vec \cite{ribeiro2017struc2vec} which is publicly available\footnote{https://github.com/leoribeiro/struc2vec}. Email network is available at Stanford SNAP dataset\footnote{https://snap.stanford.edu/data/email-Eu-core.html}. Chameleon and Squirrel networks~\cite{musae} have been collected from FAGCN \cite{bo2021beyond} which are publicly available\footnote{https://github.com/bdy9527/FAGCN}. PPI network has been collected from SuiteSparse website\footnote{https://sparse.tamu.edu/Pajek/yeast}. Cora and Citeseer networks are available in PyG datasets\footnote{https://pytorch-geometric.readthedocs.io/en/latest/modules/datasets.html}. The last accessed date for all datasets is Oct. 20, 2021.

\begin{table}[!htb]
\centering
\caption{Parameters used to run other methods. We set embedding dimension to 64 for all methods. We run all GNNs in PyTorch-Geometric framework using Adam Optimizer.}
\label{tab:parametersforothers}
\begin{tabular}{|p{8.4cm}|} 
\hline
\hspace{2.7cm}\textbf{Methods}:\textbf{Parameters}                                                                                                                                                                                            \\ 
\hline

GCN \cite{kipf2016semi}: number of convolutional layers = 2, ReLU activation function, learning rate = 0.01                                                                                                                             \\ 
\hline
GAT \cite{velivckovic2017graph}: number of convolutional layers = 2, head features = 8, ELU activation function, dropout rate = 0.6, learning rate = 0.005                                                                                      \\ 
\hline
Cluster-GCN \cite{chiang2019cluster}: number of convolutional layers = 2, number of partition = 32 and 64, ReLU activation function, dropout rate = 0.5, learning rate = 0.005                                                                                      \\ 
\hline
GDC \cite{klicpera2019diffusion}: column-wise normalization, PPR diffusion with alpha=0.05, Top-K sparsification with k = 128, number of convolutional layers = 2, ReLU activation function, learning rate = 0.01                                \\ 
\hline
GraphSAINT \cite{zeng2019graphsaint}: edge sampling approach, batch size = 128, number of steps = 5, sample coverage = 15, number of convolutional layers = 3, linear layer = 1, ReLU activation function, dropout rate = 0.2, learning rate = 0.01  \\
\hline
\end{tabular}
\end{table}

\begin{table*}[!h]
\centering
\caption{Parameters used to run different variants of MarkovGCN in PyTorch-Geometric framework (Fig. \ref{fig:ariresults}, Fig. \ref{fig:accuracyresults}, and Table \ref{tab:clustering}). We set embedding dimension to 64 for all methods. For all variants, we use Negative Log-Likelihood Loss Function and Adam Optimizer. $lrate$ - Learning rate, $droprate$ - dropout rate, $nlayers$ - the number of convolutional layers.}
\arrayrulecolor{black}
\begin{tabular}{c|c|c|c|c|c|c|c|c} 
\arrayrulecolor{black}\cline{1-1}\arrayrulecolor{black}\cline{2-9}
\multirow{2}{*}{Datasets}    & \multirow{2}{*}{MarkovGCN Variants} & \multicolumn{3}{c!{\color{black}\vrule}}{Parameters for Markov Clustering Steps} & \multicolumn{4}{c}{Parameters for Convolutional Layers}  \\ 
\cline{3-9}
                             &        - -\emph{markov\_agg}                             & Threshold ($\theta$)   & Inflation ($i$) & Row-Stochastic.                                                        & LeakyRelu & $lrate$ & $droprate$ & $nlayers$                                         \\ 
\arrayrulecolor{black}\cline{1-1}\arrayrulecolor{black}\cline{2-9}

\multirow{2}{*}{USAir}     
                             & False                          & 0.1   & 1.6     & FALSE                                                          & FALSE     & 0.01  & 0.5      & 4                                               \\ 
                             & True                          & 0.09  & 1.5     & FALSE                                                          & FALSE     & 0.01  & 0.5      & 4                                               \\ 
\hline
\multirow{2}{*}{Email}  
                             & False                          & 0.3   & 1.5     & FALSE                                                          & FALSE     & 0.01  & 0.4      & 3                                               \\ 
                             & True                          & 0.26  & 1.5     & FALSE                                                          & FALSE     & 0.01  & 0.3      & 3                                               \\ 
\hline
\multirow{2}{*}{PPI}    
                             & False                          & 0.75  & 1.7     & TRUE                                                           & FALSE     & 0.01  & 0.1      & 3                                               \\ 
                             & True                          & 0.35  & 1.8     & TRUE                                                           & FALSE     & 0.05  & 0.5      & 2                                               \\ 
\hline
\multirow{2}{*}{Chameleon} 
                             & False                          & 0.06  & 1.8     & FALSE                                                          & FALSE     & 0.01  & 0.7      & 2                                               \\ 
                             & True                          & 0.2   & 1.5     & TRUE                                                          & FALSE     & 0.01  & 0.5      & 3                                               \\ 
\hline
\multirow{2}{*}{Squirrel}  
                             & False                          & 0.2   & 1.5     & FALSE                                                          & FALSE     & 0.005 & 0.2      & 2                                               \\ 
                             & True                          & 0.05   & 1.5     & FALSE                                                          & FALSE     & 0.01  & 0.25      & 6                                               \\ 
\hline
\multirow{2}{*}{Cora}   
                             & False                          & 0.28  & 1.5     & FALSE                                                          & FALSE     & 0.01  & 0.9      & 2                                               \\ 
                             & True                          & 0.03  & 1.3     & FALSE                                                          & FALSE     & 0.01  & 0.8      & 2                                               \\ 
\hline
\multirow{2}{*}{Citeseer}    
                             & False                          & 0.11  & 1.6     & FALSE                                                          & FALSE     & 0.01  & 0.5      & 2                                               \\ 
                             & True                          & 0.11  & 1.5     & FALSE                                                          & FALSE     & 0.01  & 0.6      & 2                                               \\
\hline
\multirow{2}{*}{Actor} 
                             & False                          & 0.35 & 1.5     & FALSE                                                          & FALSE     & 0.01  & 0.5      & 3                                               \\ 
                             & True                          & 0.3 & 1.5     & FALSE                                                          & FALSE     & 0.01  & 0.5      & 3                                               \\ 
\hline
\end{tabular}
\arrayrulecolor{black}
\label{tab:markovgcnparams}
\end{table*}

\begin{table}
\centering
\caption{The contingency table for the Chameleon network. G.T. - Ground Truth, $T_i$'s are the number of true labels and $P_i$'s represent the number of predicted labels. $(i,j)$th entry denotes the number of vertices in common between $T_i$ and $P_j$. The diagonal shaded cells represent the true positives for corresponding classes. 
}

\arrayrulecolor{black}
\begin{tabular}{!{\color{black}\vrule}c!{\color{black}\vrule}c!{\color{black}\vrule}c!{\color{black}\vrule}c!{\color{black}\vrule}c!{\color{black}\vrule}c!{\color{black}\vrule}c!{\color{black}\vrule}c!{\color{black}\vrule}c!{\color{black}\vrule}c!{\color{black}\vrule}} 
\arrayrulecolor{black}\cline{1-1}\arrayrulecolor{black}\cline{2-10}
\multirow{2}{*}{G.T.} & \multicolumn{3}{c!{\color{black}\vrule}}{GCN}                                                                               & \multicolumn{3}{c!{\color{black}\vrule}}{GDC}                                                                               & \multicolumn{3}{c!{\color{black}\vrule}}{MarkovGCN}                                                                          \\ 
\cline{2-10}
                           & P1                                      & P2                                      & P3                                      & P1                                      & P2                                      & P3                                      & P1                                      & P2                                      & P3                                       \\ 
\hhline{>{\arrayrulecolor{black}}|->{\arrayrulecolor{black}}---------|}
T1                         & {\cellcolor[rgb]{0.851,0.824,0.914}}770 & 110                                     & 41                                      & {\cellcolor[rgb]{0.851,0.824,0.914}}780 & 105                                     & 36                                      & {\cellcolor[rgb]{0.851,0.824,0.914}}837 & 55                                      & 29                                       \\ 
\hline
T2                         & 75                                      & {\cellcolor[rgb]{0.851,0.824,0.914}}527 & 98                                      & 86                                      & {\cellcolor[rgb]{0.851,0.824,0.914}}543 & 71                                      & 94                                      & {\cellcolor[rgb]{0.851,0.824,0.914}}551 & 55                                       \\ 
\hline
T3                         & 56                                      & 81                                      & {\cellcolor[rgb]{0.851,0.824,0.914}}519 & 61                                      & 115                                     & {\cellcolor[rgb]{0.851,0.824,0.914}}480 & 24                                      & 78                                      & {\cellcolor[rgb]{0.851,0.824,0.914}}554  \\
\hline
\end{tabular}
\arrayrulecolor{black}
\label{tab:chameleonconfusion}
\end{table}

\textbf{Classification, Clustering and Visualization using GNNs.} We train all GNN models for 200 epochs. After each epoch, we compute the validation and test accuracy with the so far trained model. We choose the best value of the test classification accuracy for all GNN models. The graph clustering task requires labels for all vertices to avoid any bias in the experiment. Thus, we infer the labels for all vertices using the trained model which are then compared with the ground truth labels to compute performance metrics. For the visualization task, we infer the 64-dimensional embedding from the last layer
of the GNNs, then apply t-SNE to generate 2D layouts using the default parameters and plot them using $matplotlib$ package. 


\textbf{Hyper-parameters of MarkovGCN.} We analyze the hyper-parameters of MarkovGCN extensively using a grid search technique. For other MarkovGNNs, we only use different graphs in different convolutional layers where other parameters remain the same. In MarkovGCN, we analyze three parameters in the Markov process and four parameters in convolutional layers. For Markov process, we use the following strategies:
\begin{itemize}
    \item Set threshold ($\theta$) in the ranges $\{0.001 \ldots 0.009\}$, $\{0.01 \ldots 0.09\}$, and $\{0.1 \ldots 0.9\}$, and then increase by step size $0.0005$, $0.005$, and $0.05$, respectively.
    \item Set inflation ($r$) in the range $\{1.1, \ldots 3.3\}$ and increase by step $0.2$. We do not observe much variation in performance when $r > 2$.
    \item Consider either row-stochastic or column-stochastic matrices in the Markov process. By default, MarkovGCN uses column-stochastic matrices. If row-stochastic parameter is set to \emph{TRUE}, then it uses row-stochastic matrices.
\end{itemize}
We explore the parameters for convolutional layers by setting activation function as either LeakyReLU or ReLU, learning rate of Adam optimizer in the ranges $\{0.01 \ldots 0.08\}$ and $\{0.001 \ldots 0.009\}$ increasing the step size by $0.01$ and $0.001$, respectively. We set dropout rate in the range $\{0.1 \ldots 0.9\}$ increasing the step size by $0.1$ and the number of convolutional layers from 2 to 10. Section~\ref{sec:paramsensitivity} discusses the impact for some of these parameters on MarkovGCN's performance.

\begin{figure}[!h]
    \centering
    \includegraphics[width=0.75\linewidth]{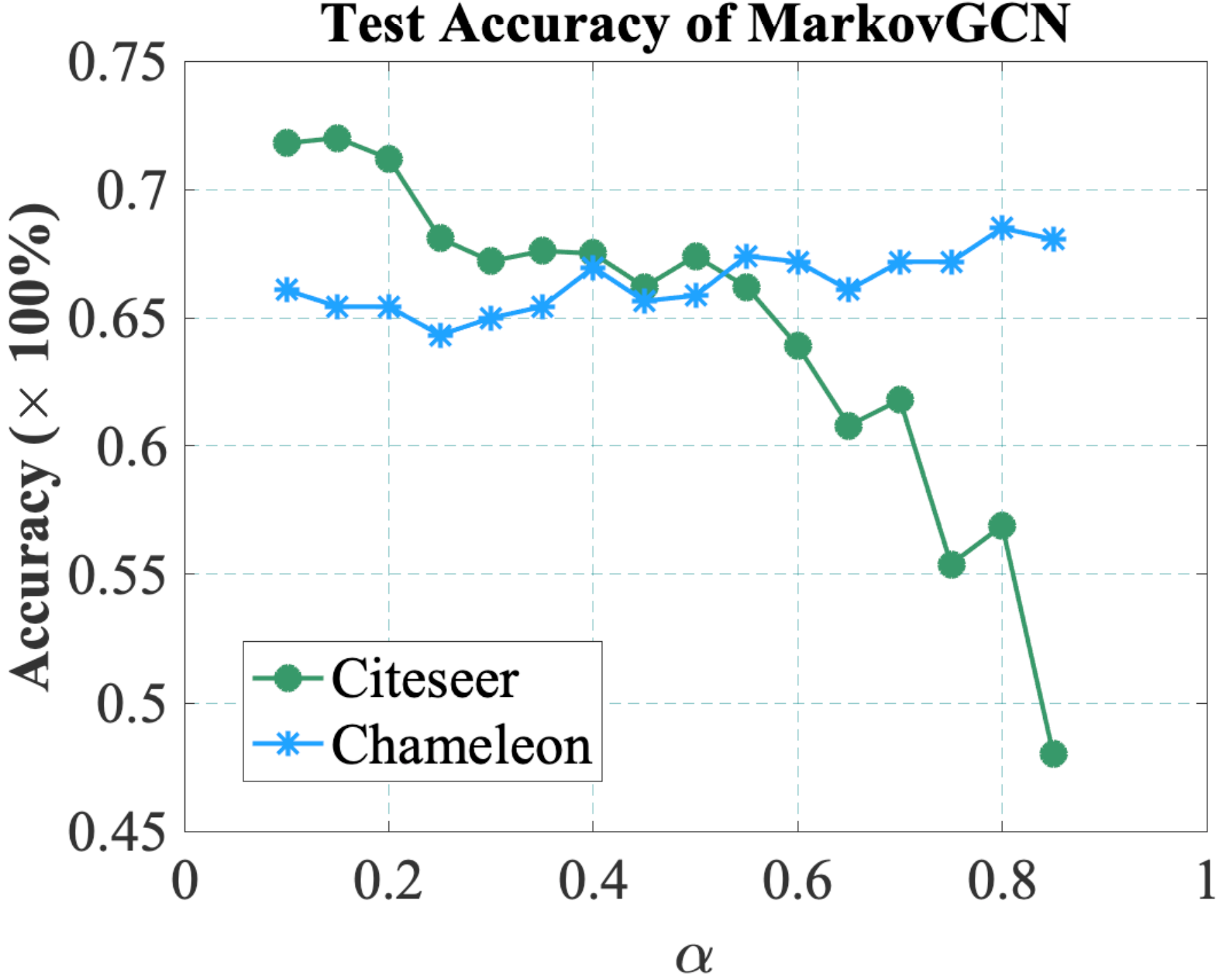}
    \caption{Sensitivity of $\alpha$ parameter for a homophilic and a heterophilic networks.}
    \label{fig:alpha_sense}
\end{figure}

\textbf{Sensitivity of $\alpha$.} In Fig. \ref{fig:alpha_sense}, we show the sensitivity of $\alpha$ parameter of Eqn. \ref{eqn:residual} using the Citeseer and Chameleon networks. We observe that the homophilic Citeseer network tends to have a lower value of $\alpha$ whereas, the heterophilic Chameleon network tends to have a higher value to produce better accuracy.

\section{Hyper Parameters}
We use PyG implementations for other methods. The I/O part is the same for all GNN methods. To run a model, we need to set some parameters. We collect those values of parameters from the papers. If no information is provided, then we use the default parameters. In Table \ref{tab:parametersforothers}, we report the parameters for other methods.

We analyze the parameter sensitivity for different variants of MarkovGCN. We use a grid search approach to search optimal value as described in Sec. \ref{sec:experimentaldetails}. After an extensive set of experiments, we pick a suitable set of parameters for the MarkovGCN models. We report the values of different parameters in Table \ref{tab:markovgcnparams} that produced better results in our experiments. These sets of parameters have been used to generate the results in Fig. \ref{fig:ariresults}, Fig. \ref{fig:accuracyresults}, and Table \ref{tab:clustering}.

\begin{table}
\centering
\caption{Adjusted Rand Index (ARI) and V-Measure for DeepWalk and struc2vec for different graphs.}\label{tab:deepwalk_struc2vec}
\arrayrulecolor{black}
\begin{tabular}{!{\color{black}\vrule}c!{\color{black}\vrule}c!{\color{black}\vrule}c!{\color{black}\vrule}c!{\color{black}\vrule}c!{\color{black}\vrule}} 
\arrayrulecolor{black}\cline{1-1}\arrayrulecolor{black}\cline{2-5}
\multirow{2}{*}{Graphs} & \multicolumn{2}{c!{\color{black}\vrule}}{ARI} & \multicolumn{2}{c!{\color{black}\vrule}}{V-Measure}  \\ 
\cline{2-5}
                        & DeepWalk & struc2vec                          & DeepWalk & struc2vec                                 \\ 
\arrayrulecolor{black}\cline{1-1}\arrayrulecolor{black}\cline{2-5}
USAir                   & 0.094    & 0.179                              & 0.150    & 0.243                                     \\ 
\hline
Email                   & 0.472    & 0.010                              & 0.704    & 0.237                                     \\ 
\hline
PPI                     & 0.030    & 0.033                              & 0.107    & 0.068                                     \\ 
\hline
Chameleon               & 0.098    & 0.035                              & 0.155    & 0.106                                     \\ 
\hline
Squirrel                & 0.013    & 0.019                              & 0.031    & 0.025                                     \\ 
\hline
Cora                    & 0.352    & 0.022                              & 0.457    & 0.101                                     \\ 
\hline
Citeseer                & 0.205    & 0.011                              & 0.212    & 0.066                                     \\
\hline
\end{tabular}
\arrayrulecolor{black}
\end{table}

\section{Additional results}
Table~\ref{tab:chameleonconfusion} shows the contingency table for the Chameleon network where the $(i,j)$th entry denotes the number of vertices in common between the ground truth class $T_i$ and the predicted class $P_j$. 
Diagonal entries in Table~\ref{tab:chameleonconfusion} denote true positive pairs of vertices stay together in both ground truth and predicted clusterings. 
These results are consistent with our hypothesis that MarkovGNN predicts the clustering patterns if available in the input graph.